\documentclass{article}

\usepackage[numbers]{natbib}
\usepackage[nonatbib, final]{stylefiles/neurips_2020}

\usepackage{graphicx}
\graphicspath{{NeuRIPS2019/figures/}}
\usepackage[utf8]{inputenc} 
\usepackage[T1]{fontenc}    
\usepackage{hyperref}       
\usepackage{url}            
\usepackage{booktabs}       
\usepackage{amsfonts}       
\usepackage{nicefrac}       
\usepackage{microtype}      
\usepackage{amsmath}
\usepackage{cancel}
\usepackage{mathtools}
\usepackage{ amssymb }
\usepackage{xcolor}
\usepackage[colorinlistoftodos]{todonotes}
\usepackage{enumitem}

\usepackage{wrapfig}
\usepackage{algorithmic}
\usepackage[ruled]{algorithm2e}
\usepackage{blindtext}
\usepackage{capt-of}

\title{Adversarial Soft Advantage Fitting:  \\Imitation Learning without Policy Optimization}

\author{%
  Paul Barde\thanks{Equal contribution. 
  \newline
 $\text{   }\quad^{\dagger}$Work conducted while interning at Ubisoft Montreal's La Forge R\&D laboratory.
 \newline
 $\text{   }\quad^\ddag$Canada CIFAR AI Chair. } $^{\,\dagger}$ \\
  Québec AI institute (Mila)\\
  McGill University\\
  \texttt{bardepau@mila.quebec} \\
  \And
  Julien Roy$^{\ast \dagger}$ \\
  Québec AI institute (Mila)\\
  Polytechnique Montréal\\
  \texttt{julien.roy@mila.quebec} \\
  \And
  Wonseok Jeon$^{\ast}$ \\
  Québec AI institute (Mila)\\
  McGill University\\
  \texttt{jeonwons@mila.quebec} \\
  \AND
  Joelle Pineau$^\ddag$ \\
  Québec AI institute (Mila)\\
  McGill University\\
  Facebook AI Research\\
  \And
  Christopher Pal$^\ddag$ \\
  Québec AI institute (Mila)\\
  Polytechnique Montréal\\
  Element AI\\
  \And
  Derek Nowrouzezahrai \\
  Québec AI institute (Mila)\\
  McGill University\\
}
\usepackage{amsthm}

\def\Figref#1{Figure~\ref{#1}}

\def\Secref#1{Section~\ref{#1}}

\def\eqref#1{equation~\ref{#1}}
\def\Eqref#1{Eq.~(\ref{#1})}

\def\Appref#1{Appendix~\ref{#1}}
\def\Thmref#1{Theorem~\ref{#1}}
\def\Lemmaref#1{Lemma~\ref{#1}}

\def\1{\bm{1}}

\DeclareMathAlphabet{\mathsfit}{\encodingdefault}{\sfdefault}{m}{sl}
\SetMathAlphabet{\mathsfit}{bold}{\encodingdefault}{\sfdefault}{bx}{n}

\def\gA{{\mathcal{A}}}

\def\gS{{\mathcal{S}}}
\def\gT{{\mathcal{T}}}

\def\sR{{\mathbb{R}}}

\newcommand{\E}{\mathbb{E}}

\newcommand{\KL}{D_{\mathrm{KL}}}
\newcommand{\JS}{D_{\mathrm{JS}}}

\DeclareMathOperator*{\argmax}{arg\,max}
\DeclareMathOperator*{\argmin}{arg\,min}

\makeatletter
\newcommand{\subalign}[1]{%
  \vcenter{%
    \Let@ \restore@math@cr \default@tag
    \baselineskip\fontdimen10 \scriptfont\tw@
    \advance\baselineskip\fontdimen12 \scriptfont\tw@
    \lineskip\thr@@\fontdimen8 \scriptfont\thr@@
    \lineskiplimit\lineskip
    \ialign{\hfil$\m@th\scriptstyle##$&$\m@th\scriptstyle{}##$\hfil\crcr
      #1\crcr
    }%
  }%
}
\makeatother

\newcommand{\pie}{\pi_{\!_E}}
\newcommand{\pig}{\pi_{\!_G}}

\newcommand{\pe}{p_{\!_E}}
\newcommand{\pg}{p_{\!_G}}

\newcommand{\pei}{p_{\!_{E}, i}}
\newcommand{\pgi}{p_{\!_{G}, i}}

\newcommand{\pigi}{\pi_{\!_{G}, i}}

\newcommand{\tildep}{\tilde{p}}
\newcommand{\tildepi}{\tilde{\pi}}
\newcommand{\tildef}{\tilde{f}}

\newtheorem{thm}{Theorem}
\newtheorem{lem}{Lemma}

\usepackage{relsize}

\newenvironment{myitemize}{
\begin{itemize}[leftmargin=*,noitemsep,topsep=0pt,parsep=0pt,partopsep=0pt]
  \setlength{\itemsep}{1pt}
  \setlength{\parskip}{1pt}
  \setlength{\parsep}{1pt}}{\end{itemize}
}

\usepackage[skip=2pt]{caption} 

\begin{document}

\maketitle

\begin{abstract}
Adversarial Imitation Learning alternates between learning a discriminator -- which tells apart expert's demonstrations from generated ones -- and a generator's policy to produce trajectories that can fool this discriminator. This alternated optimization is known to be delicate in practice since it compounds unstable adversarial training with brittle and sample-inefficient reinforcement learning. We propose to remove the burden of the policy optimization steps by leveraging a novel discriminator formulation. Specifically, our discriminator is explicitly conditioned on two policies: the one from the previous generator's iteration and a learnable policy. When optimized, this discriminator directly learns the optimal generator's policy. Consequently, our discriminator's update solves the generator's optimization problem for free: learning a policy that imitates the expert does not require an additional optimization loop. This formulation effectively cuts by half the implementation and computational burden of Adversarial Imitation Learning algorithms by removing the Reinforcement Learning phase altogether. We show on a variety of tasks that our simpler approach is competitive to prevalent Imitation Learning methods.
\end{abstract}
\section{Introduction}

Imitation Learning (IL) treats the task of learning a policy from a set of expert demonstrations. IL is effective on control problems that are challenging for traditional Reinforcement Learning (RL) methods, either due to reward function design challenges or the inherent difficulty of the task itself \cite{abbeel2004apprenticeship,ross2011reduction}. 

Most IL work can be divided into two branches: Behavioral Cloning and Inverse Reinforcement Learning. Behavioral Cloning casts IL as a supervised learning objective and seeks to imitate the expert's actions using the provided demonstrations as a fixed dataset \cite{pomerleau1991efficient}. Thus, Behavioral Cloning usually requires a lot of expert data and results in agents that struggle to generalize. As an agent deviates from the demonstrated behaviors -- straying outside the state distribution on which it was trained -- the risks of making additional errors increase, a problem known as compounding error \cite{ross2011reduction}. 

Inverse Reinforcement Learning aims to reduce compounding error by learning a reward function under which the expert policy is optimal \cite{abbeel2004apprenticeship}. Once learned, an agent can be trained (with any RL algorithm) to learn how to act at any given state of the environment. Early methods were prohibitively expensive on large environments because they required training the RL agent to convergence at each learning step of the reward function \citep{ziebart2008maximum, abbeel2004apprenticeship}. Recent approaches instead apply an adversarial formulation (Adversarial Imitation Learning, AIL) in which a discriminator learns to distinguish between expert and agent behaviors to learn the reward optimized by the expert. AIL methods allow for the use of function approximators and can in practice be used with only a few policy improvement steps for each discriminator update \citep{ho2016generative, fu2017learning, finn2016connection}.

While these advances have allowed Imitation Learning to tackle bigger and more complex environments \cite{kuefler2017imitating, ding2019goal}, they have also significantly complexified the implementation and learning dynamics of Imitation Learning algorithms. It is worth asking how much of this complexity is actually mandated. For example, in recent work, \citet{reddy2019sqil} have shown that competitive performance can be obtained by hard-coding a very simple reward function to incentivize expert-like behaviors and manage to imitate it through off-policy direct RL. \citet{reddy2019sqil} therefore remove the reward learning component of AIL and focus on the RL loop, yielding a regularized version of Behavioral Cloning. 
Motivated by these results, we also seek to simplify the AIL framework but following the opposite direction: keeping the reward learning module and removing the policy improvement loop.

We propose a simpler yet competitive AIL framework. Motivated by \citet{finn2016connection} who use the optimal discriminator form, we propose a structured discriminator that estimates the probability of demonstrated and generated behavior using a single parameterized maximum entropy policy. Discriminator learning and policy learning therefore occur simultaneously, rendering seamless generator updates: once the discriminator has been trained for a few epochs, we simply use its policy model to generate new rollouts. We call this approach Adversarial Soft Advantage Fitting (ASAF).

We make the following contributions:
\begin{myitemize}
\item \textbf{Algorithmic}: we present a novel algorithm (ASAF) designed to imitate expert demonstrations without any Reinforcement Learning step.
\item \textbf{Theoretical}: we show that our method retrieves the expert policy when trained to optimality.
\item \textbf{Empirical}: we show that ASAF outperforms prevalent IL algorithms on a variety of discrete and continuous control tasks. We also show that, in practice, ASAF can be easily modified to account for different trajectory lengths (from full length to transition-wise).
\end{myitemize}

\section{Background}
\paragraph{Markov Decision Processes (MDPs)}
We use \citet{hazan2018provably}'s notation and consider the classic $T$-horizon $\gamma$-discounted MDP $\mathcal{M}=\langle \mathcal{S}, \mathcal{A}, \mathcal{P}, \mathcal{P}_0, \gamma, r, T \rangle$. For simplicity, we assume that $\mathcal{S}$ and $\mathcal{A}$ are finite. Successor states are given by the transition distribution $\mathcal{P}(s'|s,a)\in [0,1]$, and the initial state $s_0$ is drawn from $\mathcal{P}_0(s)\in[0,1]$. Transitions are rewarded with $r(s,a)\in\mathbb{R}$ with $r$ being bounded. The discount factor and the episode horizon are $\gamma \in [0,1]$ and $T\in \mathbb{N}\cup \{\infty\}$, where $T<\infty$ for $\gamma=1$. Finally, we consider stationary stochastic policies $\pi \in \Pi : \mathcal{S}\times\mathcal{A} \rightarrow ]0,1[$ that produce trajectories $\tau = (s_0, a_0, s_1, a_1, ..., s_{T-1}, a_{T-1}, s_T)$ when executed on $\mathcal{M}$.

The probability of trajectory $\tau$ under policy $\pi$ is 
$
    P_\pi(\tau)
    \triangleq 
    \mathcal{P}_0(s_0)\prod_{t=0}^{T-1}\pi(a_t|s_t)\mathcal{P}(s_{t+1}|s_t, a_t)
$
and the corresponding marginals are defined as 
$
    d_{t, \pi}(s)
    \triangleq 
    \sum_{\tau:s_t=s}P_\pi(\tau)
$
and  
$
    d_{t, \pi}(s, a)
    \triangleq 
    \sum_{\tau:s_t=s, a_t=a}P_\pi(\tau)
    =
    d_{t, \pi}(s)\pi(a|s)
$, respectively.
With these marginals, we define the normalized discounted  state and state-action occupancy measures as
$
    d_{\pi}(s)
    \triangleq
    \frac{1}{Z(\gamma, T)}\sum_{t=0}^{T-1}\gamma^td_{t, \pi}(s)
$
and 
$
    d_{\pi}(s, a)
    \triangleq
    \frac{1}{Z(\gamma, T)}\sum_{t=0}^{T-1}\gamma^td_{t, \pi}(s, a)
    =
    d_{\pi}(s)\pi(a|s)
$
where the partition function
$Z(\gamma, T)$ is equal to $\sum_{t=0}^{T-1}\gamma^t$.
Intuitively, the state (or state-action) occupancy measure can be interpreted as the discounted visitation distribution of the states (or state-action pairs) that the agent encounters when navigating with policy $\pi$.
The expected sum of discounted rewards can be expressed in term of the occupancy measures as follows:
\begin{equation*}
\begin{aligned}
    J_{\pi}[r(s, a)] 
    \triangleq
    \E_{\tau\sim P_\pi}\left[\textstyle\sum_{t=0}^{T-1}\gamma^t \, r(s_t, a_t)\right]
    =
    Z(\gamma, T) \, \E_{(s,a)\sim d_{\pi}}[r(s, a)].
\end{aligned}
\end{equation*}
In the entropy-regularized Reinforcement Learning framework~\cite{haarnoja2018soft},
the optimal policy maximizes its entropy at each visited state in addition to the standard RL objective: 
\begin{equation*}
    \pi^*
    \triangleq
    \argmax_\pi J_\pi[r(s,a)+\alpha \mathcal{H}(\pi(\cdot|s))]\,, \quad \mathcal{H}(\pi(\cdot|s)) = \E_{a\sim \pi(\cdot|s)}[-\log(\pi(a|s))].
\end{equation*}
As shown in \cite{ziebart2010modeling, haarnoja2017reinforcement} the corresponding optimal policy is
\begin{align}
    \label{eq:max_ent_policy}
    \pi_{\mathrm{soft}}^*(a|s) =\exp \left({\alpha}^{-1} \, A^*_{\mathrm{soft}}(s,a)\right) \quad \text{with} \quad
    &A^*_{\mathrm{soft}}(s,a)
    \triangleq
    Q^*_{\mathrm{soft}}(s,a)- V^*_{\mathrm{soft}}(s),
    \\
    V^*_{\mathrm{soft}}(s)
    =
    \alpha \log\sum_{a\in\mathcal{A}}\exp \left({\alpha}^{-1} \, Q^*_{\mathrm{soft}}(s,a)\right), \,\,
    &Q^*_{\mathrm{soft}}(s,a)
    =
    r(s,a) + \gamma\mathbb{E}_{s'\sim\mathcal{P}(\cdot|s, a)}\left[ V^*_{\mathrm{soft}}(s')\right]
\end{align}

\paragraph{Maximum Causal Entropy Inverse Reinforcement Learning}
In the problem of Inverse Reinforcement Learning (IRL), it is assumed that the MDP's reward function is unknown but that demonstrations from using expert's policy $\pie$ are provided. 
Maximum causal entropy IRL \cite{ziebart2008maximum} proposes to fit a reward function $r$ from a set $\mathcal{R}$ of reward functions and retrieve the corresponding optimal policy by solving the optimization problem
\begin{equation}
\label{eq:irl_problem}
    \min_{r\in\mathcal{R}}\left(\max_{\pi} J_\pi[r(s, a) + \mathcal{H}(\pi(\cdot|s))]  \right)-J_{\pie}[r(s, a)].
\end{equation}
In brief, the problem reduces to finding a reward function $r$ for which the expert policy is optimal. In order to do so, the optimization procedure searches high entropy policies that are optimal with respect to $r$ and minimizes the difference between their returns and the return of the expert policy, eventually reaching a policy $\pi$ that approaches $\pie$. 
Most of the proposed solutions \cite{abbeel2004apprenticeship,ziebart2010modeling,ho2016generative} transpose IRL to the problem of distribution matching; \citet{abbeel2004apprenticeship} and \citet{ziebart2008maximum} used linear function approximation and proposed to match the feature expectation; \citet{ho2016generative} proposed to cast \Eqref{eq:irl_problem} with a convex reward function regularizer into the problem of minimizing the Jensen-Shannon divergence between the state-action occupancy measures:
\begin{equation}
\label{eq:irl_feature_matching}
    \min_{\pi} D_{\text{JS}}(d_\pi, d_{\pie}) - J_\pi[\mathcal{H}(\pi(\cdot|s))]
\end{equation}

\paragraph{Connections between Generative Adversarial Networks (GANs) and IRL}

For the data distribution $\pe$ and the generator distribution $\pg$ defined on the domain $\mathcal{X}$, the GAN objective~\cite{goodfellow2014generative} is
\begin{align}
    \label{eq:GAN_obj}
    \min_{\pg} \max_D L(D, \pg) \, , \quad L(D, \pg)
    \triangleq
    \mathbb{E}_{x\sim \pe}[\log D(x)]
    +
    \mathbb{E}_{x\sim \pg}[\log(1-D(x))].
\end{align}

In~\citet{goodfellow2014generative}, the maximizer of the inner problem in \Eqref{eq:GAN_obj} is shown to be
\begin{align}
\label{eq:optimal_D}
    D_{\pg}^*
    \triangleq \argmax_D L(D, \pg)
    =\frac{\pe}{\pe + \pg},
\end{align}
and the optimizer for \Eqref{eq:GAN_obj} is
$
     \argmin_{\pg}
     \max_D L(D, \pg)
     = \argmin_{\pg}L(D_{\pg}^*, \pg)=\pe
$. Later, \citet{finn2016connection} and \citet{ho2016generative}  concurrently proposed connections between GANs and IRL. The Generative Adversarial Imitation Learning (GAIL) formulation in \citet{ho2016generative} is based on matching state-action occupancy measures, while \citet{finn2016connection} considered matching trajectory distributions. Our work is inspired by the discriminator proposed and used by \citet{finn2016connection},
\begin{align}
    \label{eq:irl_D}
    D_{\theta}(\tau)
    \triangleq
    \frac{
        p_\theta(\tau)
    }{
        p_\theta(\tau)+q(\tau)
    },
\end{align}

where $p_\theta(\tau)\propto \exp r_\theta(\tau)$ with reward approximator $r_\theta$ motivated by maximum causal entropy IRL. 
Note that \Eqref{eq:irl_D} matches the form of the optimal discriminator in \Eqref{eq:optimal_D}. Although \citet{finn2016connection} do not empirically support the effectiveness of their method, the Adversarial IRL approach of \citet{fu2017learning} (AIRL) successfully used a similar discriminator for state-action occupancy measure matching.

\section{Imitation Learning without Policy Optimization}

In this section, we derive Adversarial Soft Advantage Fitting (ASAF), our novel Adversarial Imitation Learning approach. Specifically, in Section~\ref{sec:asaf_theory}, we present the theoretical foundations for ASAF to perform Imitation Learning on full-length trajectories. Intuitively, our method is based on the use of such \textit{structured discriminators} -- that match the optimal discriminator form -- to fit the trajectory distribution induced by the expert policy. This approach requires being able to evaluate and sample from the learned policy and allows us to learn that policy and train the discriminator simultaneously, thus drastically simplifying the training procedure. We present in Section~\ref{sec:policy_class} parametrization options that satisfy these requirements. Finally, in Section~\ref{sec:asaf_practical_algorithm}, we explain how to implement a practical algorithm that can be used for arbitrary trajectory-lengths, including the transition-wise case.

\subsection{Adversarial Soft Advantage Fitting -- Theoretical setting}\label{sec:asaf_theory}
Before introducing our method, we derive GAN training with a structured discriminator.

\paragraph{GAN with structured discriminator}
Suppose that we have a generator distribution $\pg$ and some arbitrary distribution $\tildep$ and that both can be evaluated efficiently, e.g., categorical distribution or probability density with normalizing flows~\cite{rezende2015variational}. 
We call a \textit{structured discriminator} a function $D_{\tildep, \pg}:\mathcal{X}\rightarrow [0,1]$ of the form
$
    D_{\tildep, \pg}(x)
    =
    {\tildep(x)}\big/({\tildep(x)+\pg(x)})
$
which matches the optimal discriminator form for~\Eqref{eq:optimal_D}. Considering our new GAN objective, we get:
\begin{align}
    \label{eq:structured_GAN_obj}
    \min_{\pg}\max_{\tildep}
    L(\tildep, \pg)\, , \quad 
    L(\tildep, \pg)
    \triangleq
    \E_{x\sim \pe}
    [\log D_{\tildep, \pg}(x)]
    +
    \E_{x\sim \pg}
    [\log (1-D_{\tildep, \pg}(x))].
\end{align}

While the unstructured discriminator $D$ from \Eqref{eq:GAN_obj} learns a mapping from $x$ to a Bernoulli distribution, we now learn a mapping from $x$ to an arbitrary distribution $\tildep$ from which we can analytically compute $D_{\tildep, \pg}(x)$. One can therefore say that $D_{\tildep, \pg}$ is \emph{parameterized} by $\tildep$. 
For the optimization problem of \Eqref{eq:structured_GAN_obj}, we have the following optima:

\begin{lem}
\label{lem:optim_D_gan}
The optimal discriminator parameter for any generator $\pg$ in \Eqref{eq:structured_GAN_obj} is equal to the expert's distribution,  
$\tildep^* 
    \triangleq \argmax_{\tildep} L(\tildep, \pg) 
    = \pe$
, and the optimal discriminator parameter is also the optimal generator, i.e.,
$$
    \pg^* 
    \triangleq \argmin_{\pg} \max_{\tildep} L(\tildep, \pg)
    = \argmin_{\pg} L(\pe, \pg)
    = \pe = \tildep^*.
$$
\end{lem}
\vspace{-15pt}
\proof{See \Appref{app:proof_of_optim_D_gan}}

Intuitively, \Lemmaref{lem:optim_D_gan} shows that the optimal discriminator parameter is also the target data distribution of our optimization problem (i.e., the optimal generator). In other words, solving the inner optimization yields the solution of the outer optimization. In practice, we update $\tildep$ to minimize the discriminator objective and use it directly as $\pg$ to sample new data.

\paragraph{Matching trajectory distributions with structured discriminator}
\label{subsec:asaf_discriminator_theory}
Motivated by the GAN with structured discriminator, we consider the trajectory distribution matching problem in IL. Here, we optimise \Eqref{eq:structured_GAN_obj} with 
$
x=\tau, 
\mathcal{X}=\gT, 
\pe=P_{\pie},
\pg=P_{\pig},
$ which yields the following objective:
\begin{align}
    \label{eq:TASAF_obj}
    \min_{\pig}\max_{\tildepi}
    L(\tildepi, \pig)
    \,,\quad L(\tildepi, \pig)
    \triangleq
    \E_{\tau\sim P_{\pie}}
    [\log D_{\tildepi, \pig}(\tau)]
    +
    \E_{\tau\sim P_{\pig}}
    [\log (1-D_{\tildepi, \pig}(\tau))],
\end{align}
with the structured discriminator:
\begin{align}
    \label{eq:simplified_trajectory_optimal_discriminator}
    D_{\tildepi, \pig}(\tau)
    = \frac{P_{\tildepi}(\tau)}{P_{\tildepi}(\tau)+P_{\pig}(\tau)}
    = \frac{q_{\tildepi}(\tau)}{q_{\tildepi}(\tau)+q_{\pig}(\tau)}.
\end{align}
Here we used the fact that $P_{\pi}(\tau)$ decomposes into two distinct products: $q_\pi(\tau)\triangleq\prod_{t=0}^{T-1}\pi(a_t|s_t)$ which depends on the stationary policy $\pi$ and $\xi(\tau)\triangleq\mathcal{P}_0(s_0)\prod_{t=0}^{T-1}\mathcal{P}(s_{t+1}|s_t, a_t)$ which accounts for the environment dynamics. Crucially, $\xi(\tau)$ cancels out in the numerator and denominator leaving $\tildepi$ as the sole parameter of this structured discriminator. In this way, $D_{\tildepi, \pig}(\tau)$ can evaluate the probability of a trajectory being generated by the expert policy simply by evaluating products of stationary policy distributions $\tildepi$ and $\pig$. With this form, we can get the following result:
\begin{thm}
\label{thm:traj_GAN}
The optimal discriminator parameter for any generator policy $\pig$ in \Eqref{eq:TASAF_obj} $\tildepi^*
\triangleq
\argmax_{\tildepi}L(\tildepi, \pig)
$ is such that
$
q_{\tildepi^*}
=
q_{\pie}
$, and using generator policy $\tildepi^*$ minimizes $L(\tildepi^*, \pig)$, i.e.,
$$
\tildepi^*
\in
\argmin_{\pig}\max_{\tildepi}L(\tildepi, \pig)
=
\argmin_{\pig}L(\tildepi^*, \pig).
$$
\end{thm}
\vspace{-15pt}
\proof{See \Appref{app:proof_of_thm_traj_GAN}}

\Thmref{thm:traj_GAN}'s benefits are similar to the ones from \Lemmaref{lem:optim_D_gan}: we can use a discriminator of the form of \Eqref{eq:simplified_trajectory_optimal_discriminator} to fit to the expert demonstrations a policy $\tildepi^*$ that simultaneously yields the optimal generator's policy and produces the same trajectory distribution as the expert policy.

\subsection{A Specific Policy Class}
\label{sec:policy_class}
The derivations of \Secref{sec:asaf_theory} rely on the use of a learnable policy that can both be evaluated and sampled from in order to fit the expert policy. A number of parameterization options that satisfy these conditions are available.

First of all, we observe that since $\pie$ is independent of $r$ and $\pi$, we can add the entropy of the expert policy $\mathcal{H}(\pie(\cdot|s))$ to the MaxEnt IRL objective of Eq.~(\ref{eq:irl_problem}) without modifying the solution to the optimization problem:
\begin{equation}
    \min_{r\in\mathcal{R}}
    \left(\max_{\pi \in \Pi} J_{\pi}[r(s,a)+\mathcal{H}(\pi(\cdot|s))]
    \right)
    - J_{\pie}[r(s,a)+\mathcal{H}(\pie(\cdot|s))]
\end{equation}
The max over policies implies that when optimising $r$, $\pi$ has already been made optimal with respect to the causal entropy augmented reward function $r'(s,a| \pi) = r(s,a) + \mathcal{H}(\pi(\cdot|s))$ and therefore it must be of the form presented in Eq.~(\ref{eq:max_ent_policy}). Moreover, since $\pi$ is optimal w.r.t. $r'$ the difference in performance $J_{\pi}[r'(s,a| \pi)]-J_{\pie}[r'(s,a|\pie)]$ is always non-negative and its minimum of 0 is only reached when $\pie$ is also optimal w.r.t. $r'$, in which case $\pie$ must also be of the form of Eq.~(\ref{eq:max_ent_policy}). 

With discrete action spaces we propose to parameterize the MaxEnt policy defined in \Eqref{eq:max_ent_policy} with the following categorical distribution $\tildepi(a|s) = \exp\left(Q_\theta(s,a) - \log\sum_{a'}\exp Q_\theta(s,a') \right)$, 
where $Q_\theta$ is a model parameterized by $\theta$ that approximates $\frac{1}{\alpha} Q^*_{\text{soft}}$.

With continuous action spaces, the soft value function involves an intractable integral over the action domain. Therefore, we approximate the MaxEnt distribution with a Normal distribution with diagonal covariance matrix like it is commonly done in the literature \cite{ haarnoja2018soft,nachum2018trustpcl}. By parameterizing the mean and variance we get a learnable density function that can be easily evaluated and sampled from. 

\subsection{Adversarial Soft Advantage Fitting (ASAF) -- practical algorithm}
\label{sec:asaf_practical_algorithm}
Section~\ref{subsec:asaf_discriminator_theory} shows that assuming $\tildepi$ can be evaluated and sampled from, we can use the structured discriminator of \Eqref{eq:simplified_trajectory_optimal_discriminator} to learn a policy $\tildepi$ that matches the expert's trajectory distribution. Section~\ref{sec:policy_class} proposes parameterizations for discrete and continuous action spaces that satisfy those assumptions.

In practice, as with GANs~\cite{goodfellow2014generative}, we do not train the discriminator to convergence as gradient-based optimisation cannot be expected to find the global optimum of non-convex problems. Instead, Adversarial Soft Advantage Fitting (ASAF) alternates between two simple steps: (1) training $D_{\tildepi, \pig}$ by minimizing the binary cross-entropy loss,
\begin{equation}
\label{eq:BCE_loss}
\begin{aligned}
    &\mathcal{L}_{BCE}(\mathcal{D}_E, \mathcal{D}_G, \tildepi) \approx -\frac{1}{n_E} \sum_{i=1}^{n_E} \log D_{\tildepi, \pig}(\tau_i^{(E)}) - \frac{1}{n_G} \sum_{i=1}^{n_G} \log \left(1 - D_{\tildepi, \pig}(\tau_i^{(G)})\right) \\
    &\text{where  }\quad \tau_i^{(E)}\sim \mathcal{D}_E \text{ ,  } \tau_i^{(G)}\sim \mathcal{D}_G \, \text{ and  } \, D_{\tildepi, \pig}(\tau) = \frac{\prod_{t=0}^{T-1}\tildepi(a_t|s_t)}{\prod_{t=0}^{T-1}\tildepi(a_t|s_t)+\prod_{t=0}^{T-1}\pig(a_t|s_t)}
\end{aligned}
\end{equation}
with minibatch sizes $n_E = n_G$, and (2) updating the generator's policy as $\pig \leftarrow \tildepi$ to minimize \Eqref{eq:TASAF_obj} (see Algorithm~\ref{alg:asaf}).

We derived ASAF considering full trajectories, yet it might be preferable in practice to split full trajectories into smaller chunks. This is particularly true in environments where trajectory length varies a lot or tends to infinity. 
 
To investigate whether the practical benefits of using partial trajectories hurt ASAF's performance, we also consider a variation, ASAF-\textit{w}, where we treat trajectory-windows of size \textit{w} as if they were full trajectories. Note that considering windows as full trajectories results in approximating that the initial state of these sub-trajectories have equal probability under the expert's and the generator's policy (this is easily seen when deriving \Eqref{eq:simplified_trajectory_optimal_discriminator}).

\begin{tabular}{@{}p{0.425\textwidth}p{0.54\textwidth}@{}}
In the limit, ASAF-1 (window-size of 1) becomes a transition-wise algorithm which can be desirable if one wants to collect rollouts asynchronously or has only access to unsequential expert data. While ASAF-1 may work well in practice it essentially assumes that the expert's and the generator's policies have the same state occupancy measure, which is incorrect until actually recovering the true expert policy.
&
\vspace{-0.25cm}
\begin{algorithm}[H]
    \begin{algorithmic}
        \caption{\label{alg:asaf}ASAF}
        \REQUIRE expert trajectories $\mathcal{D}_E = \{\tau_i\}_{i=1}^{N_E}$
        \STATE Randomly initialize $\tildepi$ and set $\pig \leftarrow \tildepi$
        \FOR{steps $m=0$ to $M$}
            \STATE Collect trajectories $\mathcal{D}_G = \{\tau_i\}_{i=1}^{N_G}$ using $\pig$
            \STATE Update $\tildepi$ by minimizing \Eqref{eq:BCE_loss}
            \STATE Set $\pig \leftarrow \tildepi$
        \ENDFOR
    \end{algorithmic}
\end{algorithm}
\end{tabular}
Finally, to offer a complete family of algorithms based on the structured discriminator approach, we show in \Appref{app:ASQF} that this assumption is not mandatory and derive a transition-wise algorithm based on Soft Q-function Fitting (rather than soft advantages) that also gets rid of the RL loop. We call this algorithm ASQF. While theoretically sound, we found that in practice, ASQF is outperformed by ASAF-1 in more complex environments (see Section~\ref{sec:results_and_discussion}).

\section{Related works}
\citet{ziebart2008maximum} first proposed MaxEnt IRL, the foundation of modern IL. \citet{ziebart2010modeling} further elaborated MaxEnt IRL as well as deriving the optimal form of the MaxEnt policy at the core of our methods. \citet{finn2016connection} proposed a GAN formulation to IRL that leveraged the energy based models of \citet{ziebart2010modeling}. \citet{finn2016guided}'s implementation of this method, however, relied on processing full trajectories with Linear Quadratic Regulator and on optimizing with guided policy search, to manage the high variance of trajectory costs. To retrieve robust rewards, \citet{fu2017learning} proposed a straightforward transposition of \cite{finn2016connection} to state-action transitions. In doing so, they had to however do away with a GAN objective during policy optimization, consequently minimizing the Kullback–Leibler divergence from the expert occupancy measure to the policy occupancy measure (instead of the Jensen-Shannon divergence)~\cite{ghasemipour2019divergence}. 

Later works \cite{sasaki2018sample, Kostrikov2020Imitation} move away from the Generative Adversarial formulation. To do so, \citet{sasaki2018sample} directly express the expectation of the Jensen-Shannon divergence between the occupancy measures in term of the agent's Q-function, which can then be used to optimize the agent's policy with off-policy Actor-Critic \cite{degris2012off}. Similarly, \citet{Kostrikov2020Imitation} use Dual Stationary Distribution Correction Estimation \cite{nachum2019dualdice} to approximate the Q-function on the expert's demonstrations before optimizing the agent's policy under the initial state distribution using the reparametrization trick \cite{haarnoja2018soft}. While \cite{sasaki2018sample,Kostrikov2020Imitation} are related to our methods in their interests in learning directly the value function, they differ in their goal and thus in the resulting algorithmic complexity. Indeed, they aim at improving the sample efficiency in terms of environment interaction and therefore move away from the algorithmically simple Generative Adversarial formulation towards more complicated divergence minimization methods. In doing so, they further complicate the Imitation Learning methods while still requiring to explicitly learn a policy. Yet, simply using the Generative Adversarial formulation with an Experience Replay Buffer can significantly improve the sample efficiency \cite{kostrikov2018discriminatoractorcritic}. 
For these reasons, and since our aim is to propose efficient yet simple methods, we focus on the Generative Adversarial formulation.

While \citet{reddy2019sqil} share our interest for simpler IL methods, they pursue an opposite approach to ours. They propose to eliminate the reward learning steps of IRL by simply hard-coding a reward of 1 for expert's transitions and of 0 for agent's transitions. They then use Soft Q-learning \cite{haarnoja2017reinforcement} to learn a value function by sampling transitions in equal proportion from the expert's and agent's buffers. Unfortunately, once the learner accurately mimics the expert, it collects expert-like transitions that are labeled with a reward of 0 since they are generated and not coming from the demonstrations. This effectively causes the reward of expert-like behavior to decay as the agent improves and can severely destabilize learning to a point where early-stopping becomes required \cite{reddy2019sqil}.

Our work builds on \cite{finn2016connection}, yet its novelty is to explicitly express the probability of a trajectory in terms of the policy in order to directly learn this latter when training the discriminator. In contrast, \cite{fu2017learning} considers a transition-wise discriminator with un-normalized probabilities which makes it closer to ASQF (Appendix~\ref{app:ASQF}) than to ASAF-1. Additionally, AIRL \cite{fu2017learning} minimizes the Kullback-Leiber Divergence \cite{ghasemipour2019divergence} between occupancy measures whereas ASAF minimizes the Jensen-Shanon Divergence between trajectory distributions.

Finally, Behavioral Cloning uses the loss function from supervised learning (classification or regression) to match expert's actions given expert's states and suffers from compounding error due to co-variate shift \cite{ross2010efficient} since its data is limited to the demonstrated state-action pairs without environment interaction. Contrarily, ASAF-1 uses the binary cross entropy loss in \Eqref{eq:BCE_loss} and does not suffer from compounding error as it learns on both generated and expert's trajectories.

\section{Results and discussion}

We evaluate our methods on a variety of discrete and continuous control tasks. Our results show that, in addition to drastically simplifying the adversarial IRL framework, our methods perform on par or better than previous approaches on all but one environment. 
When trajectory length is really long or drastically varies across episodes (see MuJoCo experiments Section~\ref{sec:mujoco_results}), we find that using sub-trajectories with fixed window-size (ASAF-\textit{w} or ASAF-1) significantly outperforms its full trajectory counterpart ASAF.

\subsection{Experimental setup}
\label{sec:results_and_discussion}
We compare our algorithms ASAF, ASAF-\textit{w} and ASAF-1 against GAIL~\cite{ho2016generative}, the predominant Adversarial Imitation Learning algorithm in the litterature, and AIRL~\cite{fu2017learning}, one of its variations that also leverages the access to the generator's policy distribution. Additionally, we compare against SQIL~\cite{reddy2019sqil}, a recent Reinforcement Learning-only approach to Imitation Learning that proved successful on high-dimensional tasks. Our implementations of GAIL and AIRL use PPO \cite{schulman2017proximal} instead of TRPO \cite{schulman2015trust} as it has been shown to improve performance \cite{kostrikov2018discriminatoractorcritic}. Finally, to be consistent with \cite{ho2016generative}, we do not use causal entropy regularization.  

For all tasks except MuJoCo, we selected the best performing hyperparameters through a random search of equal budget for each algorithm-environment pair (see Appendix~\ref{app:hyperparameters}) and the best configuration is retrained on ten random seeds. For the MuJoCo experiments, GAIL required extensive tuning (through random searches) of both its RL and IRL components to achieve satisfactory performances. Our methods, ASAF-\textit{w} and ASAF-1, on the other hand showed much more stable and robust to hyperparameterization, which is likely due to their simplicity.  
SQIL used the same SAC\cite{haarnoja2018soft} implementation  and hyperparameters that were used to generate the expert demonstrations.

Finally for each task, all algorithms use the same neural network architectures for their policy and/or discriminator (see full description in \Appref{app:hyperparameters}). 
Expert demonstrations are either generated by hand (mountaincar), using open-source bots (Pommerman) or from our implementations of SAC and PPO (all remaining). More details are given in \Appref{app:environments}.

\subsection{Experiments on classic control and Box2D tasks (discrete and continuous)}
\label{sec:classic_control_results}

\begin{figure}[h]
    \centering
    \includegraphics[width=1.\textwidth]{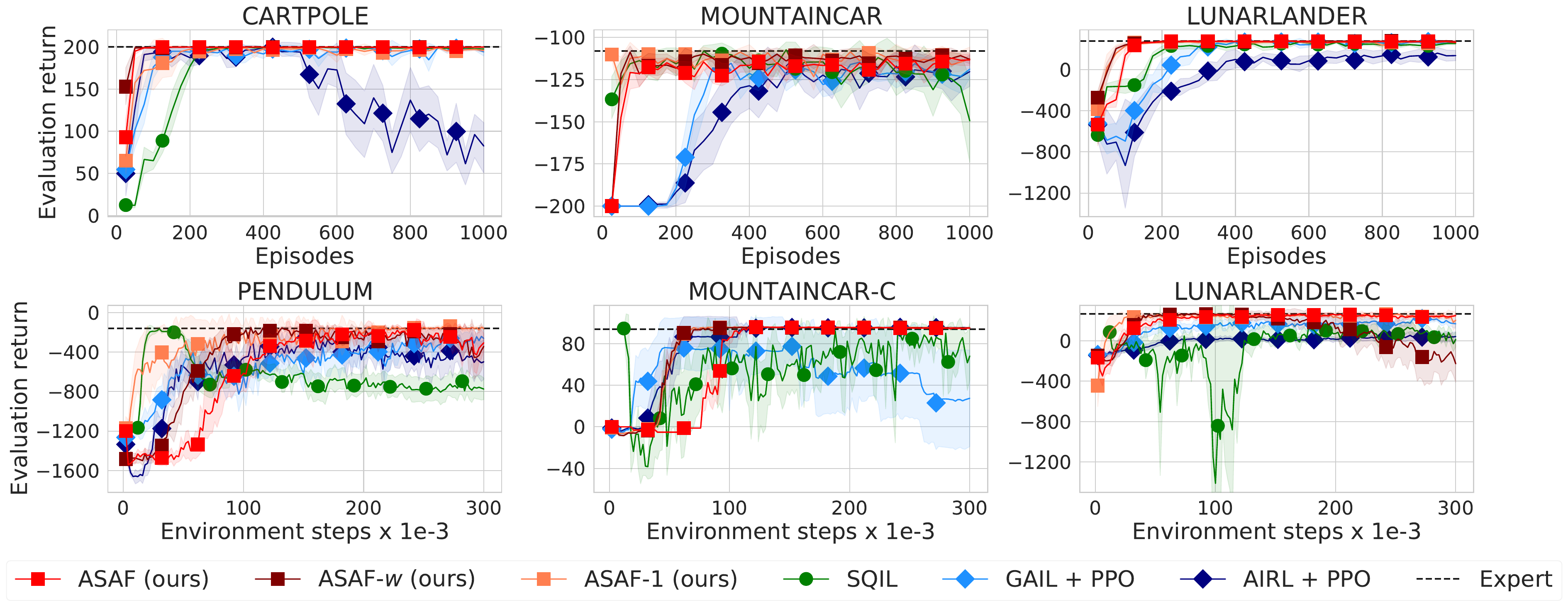}
    \caption{Results on classic control and Box2D tasks for 10 expert demonstrations. First row contains discrete actions environments, second row corresponds to continuous control.}
    \label{fig:results_toy}
\end{figure}

\Figref{fig:results_toy} shows that ASAF and its approximate variations ASAF-1 and ASAF-\textit{w} quickly converge to expert's performance (here \textit{w} was tuned to values between 32 to 200, see Appendix~\ref{app:hyperparameters} for selected window-sizes). This indicates that the practical benefits of using shorter trajectories or even just transitions does not hinder performance on these simple tasks. Note that for Box2D and classic control environments, we retrain the best configuration of each algorithm for twice as long than was done in the hyperparameter search, which allows to uncover unstable learning behaviors. \Figref{fig:results_toy} shows that our methods display much more stable learning: their performance rises until they match the expert's and does not decrease once it is reached. This is a highly desirable property for an Imitation Learning algorithm since in practice one does not have access to a reward function and thus cannot monitor the performance of the learning algorithm to trigger early-stopping. The baselines on the other hand experience occasional performance drops. For GAIL and AIRL, this is likely due to the concurrent RL and IRL loops, whereas for SQIL, it has been noted that an effective reward decay can occur when accurately mimicking the expert \cite{reddy2019sqil}. This instability is particularly severe in the continuous control case. 
In practice, all three baselines use early stopping to avoid performance decay \cite{reddy2019sqil}.

\subsection{Experiments on MuJoCo (continuous control)}
\label{sec:mujoco_results}
\begin{figure}[h]
    \centering
    \includegraphics[width=1.\textwidth]{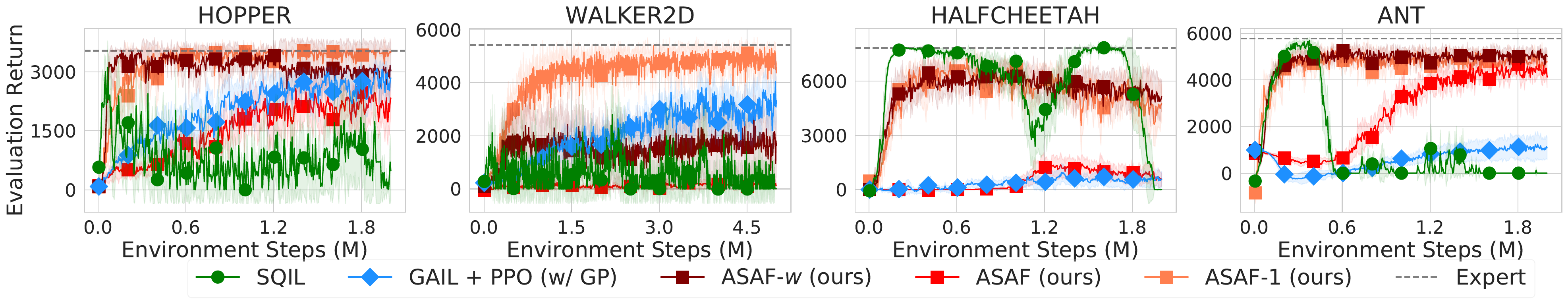}
    \caption{Results on MuJoCo tasks for 25 expert demonstrations. }
    \label{fig:mujoco_results}
\end{figure}

To scale up our evaluations in continuous control we use the popular MuJoCo benchmarks. In this domain, the trajectory length is either fixed at a large value (1000 steps on HalfCheetah) or varies a lot across episodes due to termination when the character falls down (Hopper, Walker2d and Ant). \Figref{fig:mujoco_results} shows that these trajectory characteristics hinder ASAF's learning as ASAF requires collecting multiple episodes for every update, while ASAF-1 and ASAF-\textit{w} perform well and are more sample-efficient than ASAF in these scenarios.
We focus on GAIL since \cite{fu2017learning} claim that AIRL performs on par with it on MuJoCo environments. In \Figref{fig:gail_gradient_penalty} in Appendix~\ref{app:additional_experiments} we evaluate GAIL both with and without gradient penalty (GP) on discriminator updates~\cite{gulrajani2017improved, kostrikov2018discriminatoractorcritic} and while GAIL was originally proposed without GP~\cite{ho2016generative},
we empirically found that GP prevents the discriminator to overfit and enables RL to exploit dense rewards, which highly improves its sample efficiency. Despite these ameliorations, GAIL proved to be quite inconsistent across environments despite substantial efforts on hyperparameter tuning. On the other hand, ASAF-1 performs well across all environments. Finally, we see that SQIL's instability is exacerbated on MuJoCo.

\subsection{Experiments on Pommerman (discrete control)}
\label{sec:pommerman_results}

\begin{figure}[h]
    \centering
    \begin{tabular}{ccc}
        \includegraphics[width=.2\textwidth]{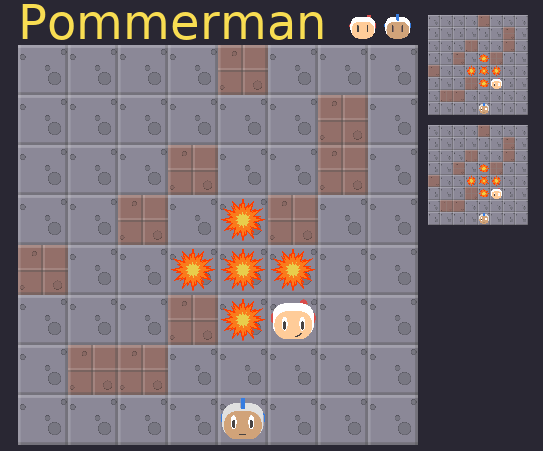} &
        \includegraphics[width=.45\textwidth]{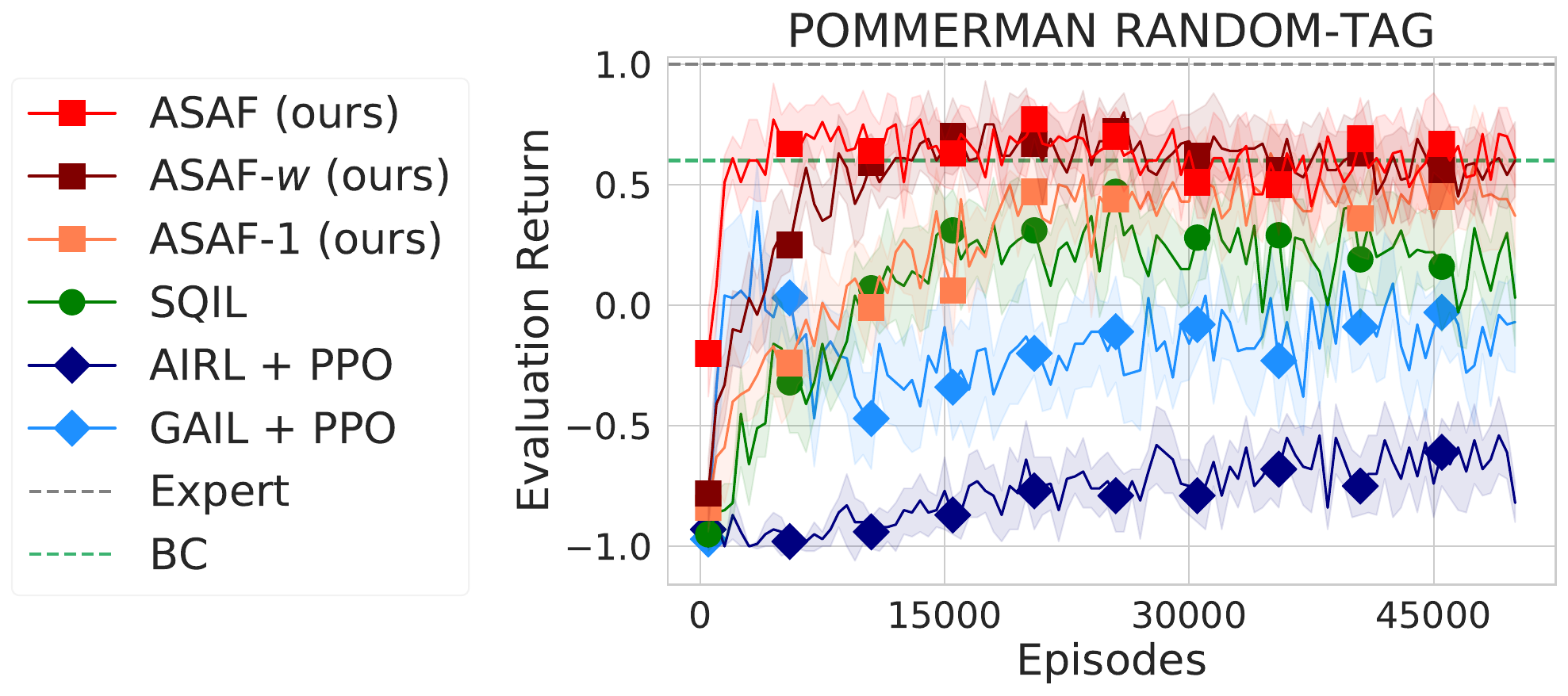} &
        \includegraphics[width=.26\textwidth]{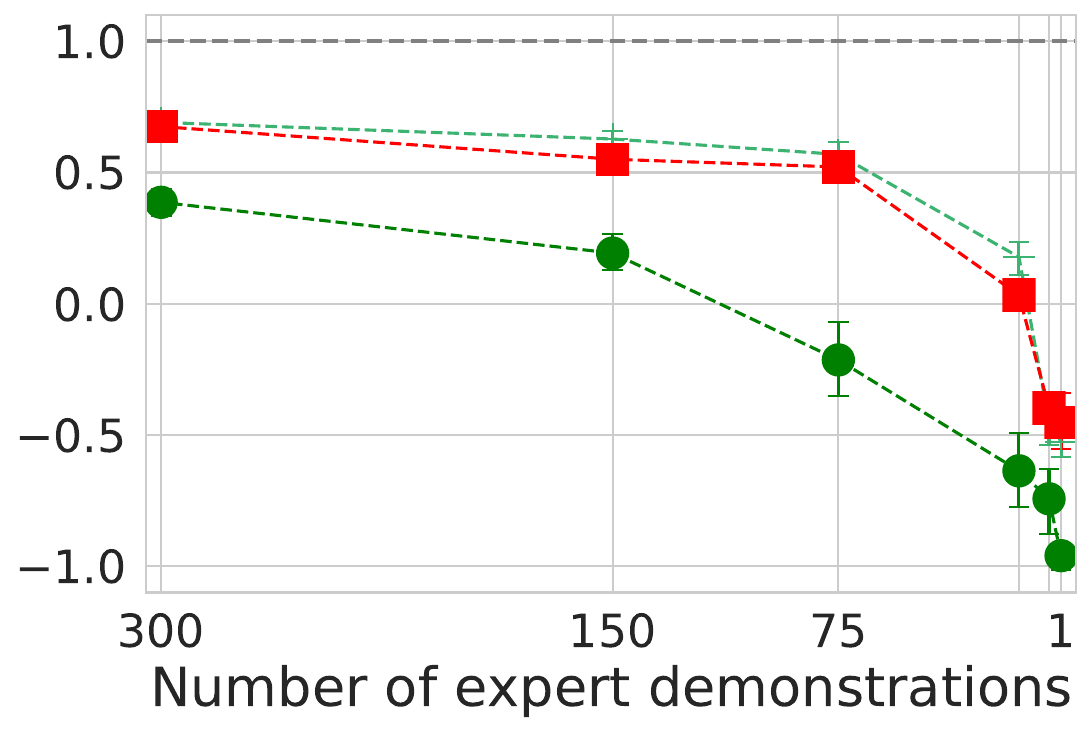} 
    \end{tabular}
    \caption{Results on Pommerman Random-Tag: (Left) Snapshot of the environment. (Center) Learning measured as evaluation return over episodes for 150 expert trajectories  (Right) Average return on last 20\% of training for decreasing number of expert trajectories [300, 150, 75, 15, 5, 1].}
    \label{fig:pommerman_results_randomTag}
\end{figure}

Finally, to scale up our evaluations in discrete control environments, we consider the domain of Pommerman~\cite{resnick2018pommerman}, a challenging and very dynamic discrete control environment that uses rich and high-dimensional observation spaces (see Appendix~\ref{app:environments}). We perform evaluations of all of our methods and baselines on a 1 vs 1 task where a learning agent plays against a random agent, the opponent. The goal for the learning agent is to navigate to the opponent and eliminate it using expert demonstrations provided by the champion algorithm of the FFA 2018 competition~\cite{zhou2018hybrid}. We removed the ability of the opponent to lay bombs so that it doesn't accidentally eliminate itself. Since it can still move around, it is however surprisingly tricky to eliminate: the expert has to navigate across the whole map, lay a bomb next to the opponent and retreat to avoid eliminating itself. This entire routine has then to be repeated several times until finally succeeding since the opponent will often avoid the hit by chance. We refer to this task as \textit{Pommerman Random-Tag}. Note that since we measure success of the imitation task with the win-tie-lose outcome (sparse performance metric), a learning agent has to truly reproduce the expert behavior until the very end of trajectories to achieve higher scores. Figure~\ref{fig:pommerman_results_randomTag} shows that all three variations of ASAF as well as Behavioral Cloning (BC) outperform the baselines.

\section{Conclusion}

We propose an important simplification to the Adversarial Imitation Learning framework by removing the Reinforcement Learning optimisation loop altogether. We show that, by using a particular form for the discriminator, our method recovers a policy that matches the expert's trajectory distribution. We evaluate our approach against prior works on many different benchmarking tasks and show that our method (ASAF) compares favorably to the predominant Imitation Learning algorithms. The approximate versions, ASAF-\textit{w} and ASAF-1, that use sub-trajectories yield a flexible algorithms that work well both on short and long time horizons. Finally, our approach still involves a reward learning module through its discriminator, and it would be interesting in future work to explore how ASAF can be used to learn robust rewards, along the lines of~\citet{fu2017learning}.

\section*{Broader Impact}
Our contributions are mainly theoretical and aim at simplifying current Imitation Learning methods. We do not propose new applications nor use sensitive data or simulator. Yet our method can ease and promote the use, design and development of Imitation Learning algorithms and may eventually lead to applications outside of simple and controlled simulators. We do not pretend to discuss the ethical implications of the general use of autonomous agents but we rather try to investigate what are some of the differences in using Imitation Learning rather than reward oriented methods in the design of such agents.

Using only a scalar reward function to specify the desired behavior of an autonomous agent is a challenging task as one must weight different desiderata and account for unsuspected behaviors and situations. Indeed, it is well known in practice that Reinforcement Learning agents tend to find bizarre ways of exploiting the reward signal without solving the desired task. The fact that it is difficult to specify and control the behavior or an RL agents is a major flaw that prevent current methods to be applied to risk sensitive situations. On the other hand, Imitation Learning proposes a more natural way of specifying nuanced preferences by demonstrating desirable ways of solving a task. Yet, IL also has its drawbacks. First of all one needs to be able to demonstrate the desired behavior and current methods tend to be only as good as the demonstrator. Second, it is a challenging problem to ensure that the agent will be able to adapt to new situations that do not resemble the demonstrations. For these reasons, it is clear for us that additional safeguards are required in order to apply Imitation Learning (and Reinforcement Learning) methods to any application that could effectively have a real world impact. 

\begin{ack}
We thank Eloi Alonso, Olivier Delalleau, Félix G. Harvey, Maxim Peter and the entire research team at Ubisoft Montreal's La Forge R\&D laboratory. Their feedback and comments contributed significantly to this work.
Christopher Pal and Derek Nowrouzezahrai acknowledge funding from the
Fonds de Recherche Nature et
Technologies (FRQNT), Ubisoft Montreal and Mitacs’ Accelerate Program in
support of our work,
as well as Compute Canada for providing computing resources. Derek and Paul also acknowledge support from the NSERC Industrial Research Chair program.
\end{ack}
\bibliography{sources.bib}

\begin{thebibliography}{30}
\providecommand{\natexlab}[1]{#1}
\providecommand{\url}[1]{\texttt{#1}}
\expandafter\ifx\csname urlstyle\endcsname\relax
  \providecommand{\doi}[1]{doi: #1}\else
  \providecommand{\doi}{doi: \begingroup \urlstyle{rm}\Url}\fi

\bibitem[Abbeel and Ng(2004)]{abbeel2004apprenticeship}
Pieter Abbeel and Andrew~Y Ng.
\newblock Apprenticeship learning via inverse reinforcement learning.
\newblock In \emph{Proceedings of the 21st International Conference on Machine
  Learning (ICML)}, 2004.

\bibitem[Degris et~al.(2012)Degris, White, and Sutton]{degris2012off}
Thomas Degris, Martha White, and Richard~S Sutton.
\newblock Off-policy actor-critic.
\newblock In \emph{Proceedings of the 29th International Conference on Machine
  Learning (ICML)}, pages 179--186, 2012.

\bibitem[Ding et~al.(2019)Ding, Florensa, Abbeel, and Phielipp]{ding2019goal}
Yiming Ding, Carlos Florensa, Pieter Abbeel, and Mariano Phielipp.
\newblock Goal-conditioned imitation learning.
\newblock In \emph{Advances in Neural Information Processing Systems
  (NeurIPS)}, pages 15298--15309, 2019.

\bibitem[Finn et~al.(2016{\natexlab{a}})Finn, Christiano, Abbeel, and
  Levine]{finn2016connection}
Chelsea Finn, Paul Christiano, Pieter Abbeel, and Sergey Levine.
\newblock A connection between generative adversarial networks, inverse
  reinforcement learning, and energy-based models.
\newblock \emph{arXiv preprint arXiv:1611.03852}, 2016{\natexlab{a}}.

\bibitem[Finn et~al.(2016{\natexlab{b}})Finn, Levine, and
  Abbeel]{finn2016guided}
Chelsea Finn, Sergey Levine, and Pieter Abbeel.
\newblock Guided cost learning: Deep inverse optimal control via policy
  optimization.
\newblock In \emph{Proceedings of the 33rd International Conference on Machine
  Learning (ICML)}, pages 49--58, 2016{\natexlab{b}}.

\bibitem[Fu et~al.(2017)Fu, Luo, and Levine]{fu2017learning}
Justin Fu, Katie Luo, and Sergey Levine.
\newblock Learning robust rewards with adversarial inverse reinforcement
  learning.
\newblock In \emph{Proceedings of the 5th International Conference on Learning
  Representations (ICLR)}, 2017.

\bibitem[Ghasemipour et~al.(2019)Ghasemipour, Zemel, and
  Gu]{ghasemipour2019divergence}
Seyed Kamyar~Seyed Ghasemipour, Richard Zemel, and Shixiang Gu.
\newblock A divergence minimization perspective on imitation learning methods.
\newblock In \emph{Proceedings of the 3rd Conference on Robot Learning (CoRL)},
  2019.

\bibitem[Goodfellow et~al.(2014)Goodfellow, Pouget-Abadie, Mirza, Xu,
  Warde-Farley, Ozair, Courville, and Bengio]{goodfellow2014generative}
Ian Goodfellow, Jean Pouget-Abadie, Mehdi Mirza, Bing Xu, David Warde-Farley,
  Sherjil Ozair, Aaron Courville, and Yoshua Bengio.
\newblock Generative adversarial nets.
\newblock In \emph{Advances in Neural Information Processing Systems
  (NeurIPS)}, pages 2672--2680, 2014.

\bibitem[Gulrajani et~al.(2017)Gulrajani, Ahmed, Arjovsky, Dumoulin, and
  Courville]{gulrajani2017improved}
Ishaan Gulrajani, Faruk Ahmed, Martin Arjovsky, Vincent Dumoulin, and Aaron~C
  Courville.
\newblock Improved training of {W}asserstein {GAN}s.
\newblock In \emph{Advances in Neural Information Processing Systems
  (NeurIPS)}, pages 5767--5777, 2017.

\bibitem[Haarnoja et~al.(2017)Haarnoja, Tang, Abbeel, and
  Levine]{haarnoja2017reinforcement}
Tuomas Haarnoja, Haoran Tang, Pieter Abbeel, and Sergey Levine.
\newblock Reinforcement learning with deep energy-based policies.
\newblock In \emph{Proceedings of the 34th International Conference on Machine
  Learning (ICML)}, pages 1352--1361, 2017.

\bibitem[Haarnoja et~al.(2018)Haarnoja, Zhou, Hartikainen, Tucker, Ha, Tan,
  Kumar, Zhu, Gupta, Abbeel, et~al.]{haarnoja2018soft}
Tuomas Haarnoja, Aurick Zhou, Kristian Hartikainen, George Tucker, Sehoon Ha,
  Jie Tan, Vikash Kumar, Henry Zhu, Abhishek Gupta, Pieter Abbeel, et~al.
\newblock Soft actor-critic algorithms and applications.
\newblock \emph{arXiv preprint arXiv:1812.05905}, 2018.

\bibitem[Hazan et~al.(2018)Hazan, Kakade, Singh, and
  Van~Soest]{hazan2018provably}
Elad Hazan, Sham~M Kakade, Karan Singh, and Abby Van~Soest.
\newblock Provably efficient maximum entropy exploration.
\newblock \emph{arXiv preprint arXiv:1812.02690}, 2018.

\bibitem[Ho and Ermon(2016)]{ho2016generative}
Jonathan Ho and Stefano Ermon.
\newblock Generative adversarial imitation learning.
\newblock In \emph{Advances in Neural Information Processing Systems
  (NeurIPS)}, pages 4565--4573, 2016.

\bibitem[Kostrikov et~al.(2019)Kostrikov, Agrawal, Dwibedi, Levine, and
  Tompson]{kostrikov2018discriminatoractorcritic}
Ilya Kostrikov, Kumar~Krishna Agrawal, Debidatta Dwibedi, Sergey Levine, and
  Jonathan Tompson.
\newblock Discriminator-actor-critic: Addressing sample inefficiency and reward
  bias in adversarial imitation learning.
\newblock In \emph{Proceedings of the 7th International Conference on Learning
  Representations (ICLR)}, 2019.

\bibitem[Kostrikov et~al.(2020)Kostrikov, Nachum, and
  Tompson]{Kostrikov2020Imitation}
Ilya Kostrikov, Ofir Nachum, and Jonathan Tompson.
\newblock Imitation learning via off-policy distribution matching.
\newblock In \emph{Proceedings of the 8th International Conference on Learning
  Representations (ICLR)}, 2020.

\bibitem[Kuefler et~al.(2017)Kuefler, Morton, Wheeler, and
  Kochenderfer]{kuefler2017imitating}
Alex Kuefler, Jeremy Morton, Tim Wheeler, and Mykel Kochenderfer.
\newblock Imitating driver behavior with generative adversarial networks.
\newblock In \emph{Proceedings of 2017 IEEE Intelligent Vehicles Symposium
  (IV)}, pages 204--211, 2017.

\bibitem[Nachum et~al.(2018)Nachum, Norouzi, Xu, and
  Schuurmans]{nachum2018trustpcl}
Ofir Nachum, Mohammad Norouzi, Kelvin Xu, and Dale Schuurmans.
\newblock Trust-{PCL}: An off-policy trust region method for continuous
  control.
\newblock In \emph{Proceedings of the 6th International Conference on Learning
  Representations (ICLR)}, 2018.

\bibitem[Nachum et~al.(2019)Nachum, Chow, Dai, and Li]{nachum2019dualdice}
Ofir Nachum, Yinlam Chow, Bo~Dai, and Lihong Li.
\newblock Dual{DICE}: Behavior-agnostic estimation of discounted stationary
  distribution corrections.
\newblock In \emph{Advances in Neural Information Processing Systems
  (NeurIPS)}, pages 2318--2328, 2019.

\bibitem[Pomerleau(1991)]{pomerleau1991efficient}
Dean~A Pomerleau.
\newblock Efficient training of artificial neural networks for autonomous
  navigation.
\newblock \emph{Neural computation}, 3\penalty0 (1):\penalty0 88--97, 1991.

\bibitem[Reddy et~al.(2019)Reddy, Dragan, and Levine]{reddy2019sqil}
Siddharth Reddy, Anca~D. Dragan, and Sergey Levine.
\newblock {SQIL}: Imitation learning via reinforcement learning with sparse
  rewards, 2019.

\bibitem[Resnick et~al.(2018)Resnick, Eldridge, Ha, Britz, Foerster, Togelius,
  Cho, and Bruna]{resnick2018pommerman}
Cinjon Resnick, Wes Eldridge, David Ha, Denny Britz, Jakob Foerster, Julian
  Togelius, Kyunghyun Cho, and Joan Bruna.
\newblock Pommerman: A multi-agent playground.
\newblock \emph{arXiv preprint arXiv:1809.07124}, 2018.

\bibitem[Rezende and Mohamed(2015)]{rezende2015variational}
Danilo Rezende and Shakir Mohamed.
\newblock Variational inference with normalizing flows.
\newblock In \emph{Proceedings of the 32nd International Conference on Machine
  Learning (ICML)}, pages 1530--1538, 2015.

\bibitem[Ross and Bagnell(2010)]{ross2010efficient}
St{\'e}phane Ross and Drew Bagnell.
\newblock Efficient reductions for imitation learning.
\newblock In \emph{Proceedings of the 13th International Conference on
  Artificial Intelligence and Statistics (AISTATS)}, pages 661--668, 2010.

\bibitem[Ross et~al.(2011)Ross, Gordon, and Bagnell]{ross2011reduction}
St{\'e}phane Ross, Geoffrey Gordon, and Drew Bagnell.
\newblock A reduction of imitation learning and structured prediction to
  no-regret online learning.
\newblock In \emph{Proceedings of the 14th International Conference on
  Artificial Intelligence and Statistics (AISTATS)}, pages 627--635, 2011.

\bibitem[Sasaki et~al.(2018)Sasaki, Yohira, and Kawaguchi]{sasaki2018sample}
Fumihiro Sasaki, Tetsuya Yohira, and Atsuo Kawaguchi.
\newblock Sample efficient imitation learning for continuous control.
\newblock In \emph{Proceedings of the 6th International Conference on Learning
  Representations (ICLR)}, 2018.

\bibitem[Schulman et~al.(2015)Schulman, Levine, Abbeel, Jordan, and
  Moritz]{schulman2015trust}
John Schulman, Sergey Levine, Pieter Abbeel, Michael Jordan, and Philipp
  Moritz.
\newblock Trust region policy optimization.
\newblock In \emph{Proceedings of the 32nd International Conference on Machine
  Learning (ICML)}, pages 1889--1897, 2015.

\bibitem[Schulman et~al.(2017)Schulman, Wolski, Dhariwal, Radford, and
  Klimov]{schulman2017proximal}
John Schulman, Filip Wolski, Prafulla Dhariwal, Alec Radford, and Oleg Klimov.
\newblock Proximal policy optimization algorithms.
\newblock \emph{arXiv preprint arXiv:1707.06347}, 2017.

\bibitem[Zhou et~al.(2018)Zhou, Gong, Mugrai, Khalifa, Nealen, and
  Togelius]{zhou2018hybrid}
Hongwei Zhou, Yichen Gong, Luvneesh Mugrai, Ahmed Khalifa, Andy Nealen, and
  Julian Togelius.
\newblock A hybrid search agent in pommerman.
\newblock In \emph{Proceedings of the 13th International Conference on the
  Foundations of Digital Games (FDG)}, pages 1--4, 2018.

\bibitem[Ziebart(2010)]{ziebart2010modeling}
Brian~D Ziebart.
\newblock \emph{Modeling purposeful adaptive behavior with the principle of
  maximum causal entropy}.
\newblock PhD thesis, 2010.

\bibitem[Ziebart et~al.(2008)Ziebart, Maas, Bagnell, and
  Dey]{ziebart2008maximum}
Brian~D Ziebart, Andrew~L Maas, J~Andrew Bagnell, and Anind~K Dey.
\newblock Maximum entropy inverse reinforcement learning.
\newblock In \emph{Proceedings of the 23rd AAAI Conference on Artificial
  Intelligence}, pages 1433--1438, 2008.

\end{thebibliography}
\newpage
\section*{Appendix}
\appendix
\section{Proofs}
\subsection{Proof of \Lemmaref{lem:optim_D_gan}}
\label{app:proof_of_optim_D_gan}
\begin{proof}
\Lemmaref{lem:optim_D_gan} states that given $L(\tildep, \pg)$ defined in~\Eqref{eq:structured_GAN_obj}:
\begin{enumerate}[label=(\alph*)]
    \item $\displaystyle \tildep^* \triangleq \argmax_{\tildep} L(\tildep, \pg) = \pe$
    \item $\displaystyle \argmin_{\pg} L(\pe, \pg) = \pe$
\end{enumerate}

Starting with (a), we have:
\begin{align*}
    \argmax_{\tildep} L(\tildep, \pg)
    &= \argmax_{\tildep}
    \sum_{x_i} \pe(x_i)\log D_{\tildep, \pg}(x_i) + \pg(x_i)\log (1 - D_{\tildep, \pg}(x_i))\\
    &\triangleq \argmax_{\tildep} \sum_{x_i} L_i
\end{align*}
Assuming infinite discriminator's capacity, $L_i$ can be made independent for all $x_i \in \mathcal{X}$ and we can construct our optimal discriminator $D_{\tildep, \pg}^*$ as a look-up table $D_{\tildep, \pg}^*:\mathcal{X} \rightarrow \, ]0,1[ \, ; \, x_i \mapsto D^*_i$ with $D^*_i$ the optimal discriminator for each $x_i$ defined as:
\begin{equation}
    D^*_i = \argmax_{D_i}L_i = \argmax_{D_i} \pei
    \log D_i + \pgi\log (1 - D_i),
\end{equation}
with $\pgi\triangleq\pg(x_i)$, $\pei\triangleq\pe(x_i)$ and $D_i\triangleq D(x_i)$.

Recall that $D_i \in \, ]0,1[$ and that $\pgi \in \, ]0,1[$. Therefore the function $ \tildep_i \mapsto D_i = \dfrac{\tildep_i}{\tildep_i + \pgi}$ is defined for $\tildep_i \in ]0, +\infty[$. Since it is strictly monotonic over that domain we have that:
\begin{align}
D_i^*=\argmax_{D_i} L_i \, \Leftrightarrow \, \tildep^*_i =\argmax_{\tildep_i} L_i
\end{align}
Taking the derivative and setting to zero, we get:
\begin{align}
    \left.\frac{d L_i}{d \tildep_i}\right|_{\tildep_i} = 0 \, \Leftrightarrow& \,
    \tildep_i = \pei
\end{align}
The second derivative test confirms that we have a maximum, i.e. $\left.\dfrac{d^2 L_i}{d \tildep_i^2}\right|_{\tildep_i^*} < 0$. The values of $L_i$ at the boundaries of the domain of definition of $\tildep_i$ tend to $-\infty$, therefore $L_i(\tildep_i^*=\pei)$ is the global maximum of $L_i$ w.r.t. $\tildep_i$. Finally, the optimal global discriminator is given by:
\begin{equation}
\label{eq:D_optim_GAN_proof}
    D_{\tildep, \pg}^*(x) = \frac{\pe(x)}{\pe(x)+\pg(x)} \quad \forall x \in \mathcal{X}
\end{equation}
This concludes the proof for (a).

The proof for (b) can be found in the work of \citet{goodfellow2014generative}. We reproduce it here for completion. Since from (a) we know that $\tildep^*(x) = \pe(x) \, \forall x \in \mathcal{X}$, we can write the GAN objective for the optimal discriminator as:
\begin{align}
    \argmin_{\pg} L(\tildep^*, \pg)
    &= \argmin_{\pg} L(\pe, \pg) \\
    &= \argmin_{\pg} \E_{x\sim \pe}\left[\log \frac{\pe(x)}{\pe(x)+\pg(x)}\right] + \E_{x\sim \pg}\left[\log \frac{\pg(x)}{\pe(x)+\pg(x)}\right]\label{eq:L_Dopt}
\end{align}
Note that:
\begin{equation}
\label{eq:minuslog4}
    \log4 = \E_{x\sim \pe}\left[\log 2\right] + \E_{x\sim \pg}\left[\log 2\right]
\end{equation}
Adding \Eqref{eq:minuslog4} to \Eqref{eq:L_Dopt} and subtracting $\log4$ on both sides:
\begin{align}
    \argmin_{\pg} L(\pe, \pg) &= -\log4 + \E_{x\sim \pe}\left[\log \frac{2\pe(x)}{\pe(x)+\pg(x)}\right] + \E_{x \sim \pg}\left[\log \frac{2\pg(x)}{\pe(x)+\pg(x)}\right]\\
    &= -\log4 + \KL\left(\pe\left\|\frac{\pe + \pg}{2}\right.\right)+ \KL\left(\pe\left\|\frac{\pe + \pg}{2}\right.\right)\\
    &= -\log4 + 2\JS\left(\pe\left\|\pg\right.\right)
\end{align}
Where $\KL$ and $\JS$ are respectively the Kullback-Leibler and the Jensen-Shannon divergences. Since the Jensen-Shannon divergence between two distributions is always non-negative and zero if and only if the two distributions are equal, we have that $\displaystyle \argmin_{\pg} L(\pe, \pg) = \pe$.

This concludes the proof for (b). 
\end{proof}

\subsection{Proof of \Thmref{thm:traj_GAN}}
\label{app:proof_of_thm_traj_GAN}
\begin{proof}
\Thmref{thm:traj_GAN} states that given $L(\tildepi, \pig)$ defined in~\Eqref{eq:TASAF_obj}:
\begin{enumerate}[label=(\alph*)]
    \item $\displaystyle \tildepi^* \triangleq \argmax_{\tildepi} L(\tildepi, \pig) \text{ satisfies } q_{\tildepi^*} = q_{\pie}$
    \item $\displaystyle \pig^* = \tildepi^* \in \argmin_{\pig}L(\tildepi^*, \pig)$
\end{enumerate}

The proof of (a) is very similar to the one from \Lemmaref{lem:optim_D_gan}. Starting from \Eqref{eq:TASAF_obj} we have:
\begin{align}
    \argmax_{\tildepi}
    L(\tildepi, \pig)
    &=
    \argmax_{\tildepi} \sum_{\tau_i} P_{\pie}(\tau_i) \log D_{\tildepi, \pig}(\tau_i)
    +
    P_{\pig}(\tau_i) \log (1-D_{\tildepi, \pig}(\tau_i)) \\
    &=
    \argmax_{\tildepi} \sum_{\tau_i} \xi(\tau_i) \left( q_{\pie}(\tau_i) \log D_{\tildepi, \pig}(\tau_i)
    +
    q_{\pig}(\tau_i) \log (1-D_{\tildepi, \pig}(\tau_i))
    \right) \\
    &=
    \argmax_{\tildepi} \sum_{\tau_i} L_i
\end{align}
Like for \Lemmaref{lem:optim_D_gan}, we can optimise for each $L_i$ individually. When doing so, $\xi(\tau_i)$ can be omitted as it is constant w.r.t $\tildepi$. The rest of the proof is identical to the one of but \Lemmaref{lem:optim_D_gan} with $\pe = q_{\pie}$ and $\pg = q_{\pig}$. It follows that the max of $L(\tildepi, \pig)$ is reached for $q_{\tildepi}^* = q_{\pie}$. From that we obtain that the policy $\tildepi^*$ that makes the discriminator $D_{\tildepi^*, \pig}$ optimal w.r.t $L(\tildepi, \pig)$ is such that $q_{\tildepi^*} = q_{\tildepi}^* = q_{\pie}$ i.e. $\prod_{t=0}^{T-1}\tildepi^*(a_t|s_t) = \prod_{t=0}^{T-1}\pie(a_t|s_t) \, \forall \, \tau$. 

The proof for (b) stems from the observation that choosing $\pig = \tildepi^*$ (the policy recovered by the optimal discriminator $D_{\tildepi^*, \pig}$) minimizes $L(\tildepi^*, \pig)$:
\begin{align}
    \pig(a|s) = \tildepi^*(a|s) \,
    \, \forall \, (s,a) \in \gS\times\gA \quad
    &\Rightarrow \quad \prod_{t=0}^{T-1}\pig(a_t|s_t) = \prod_{t=0}^{T-1}\tildepi^*(a_t|s_t) \,\, \forall \tau \in \gT \\
    &\Rightarrow \quad q_{\pig}(\tau) = q_{\pie}(\tau) \,\, \forall \, \tau \in \gT \\
    &\Rightarrow \quad D_{\tildepi^*, \tildepi^*} = \frac{1}{2} \,\, \forall \,\tau \in \gT \\
    &\Rightarrow \quad L(\tildepi^*, \tildepi^*) = -\log 4
\end{align}
By multiplying the numerator and denominator of $D_{\tildepi^*, \tildepi^*}$ by $\xi(\tau)$ it can be shown in exactly the same way as in Appendix~\ref{app:proof_of_optim_D_gan} that $-\log4$ is the global minimum of $L(\tildepi^*, \pig)$.
\end{proof}

\section{Adversarial Soft Q-Fitting: transition-wise Imitation Learning without Policy Optimization}
\label{app:ASQF}
In this section we present Adversarial Soft Q-Fitting (ASQF), a principled approach to Imitation Learning without Reinforcement Learning that relies exclusively on transitions. Using transitions rather than trajectories presents several practical benefits such as the possibility to deal with asynchronously collected data or non-sequential experts demonstrations. We first present the theoretical setting for ASQF and then test it on a variety of discrete control tasks. We show that while it is theoretically sound, ASQF is often outperformed by ASAF-1, an approximation to ASAF that also allows to rely on transitions instead of trajectories.

\paragraph{Theoretical Setting}
We consider the GAN objective of \Eqref{eq:GAN_obj} with 
$x=(s,a)$, $
\mathcal{X}=\gS \times \gA$, $\pe=d_{\pie}$, $\quad
\pg=d_{\pig}$ and a discriminator $D_{\tildef, \pig}$ of the form of \citet{fu2017learning}:
\begin{equation}
\begin{aligned}
    \label{eq:ASQF_obj}
    \min_{\pig}\max_{\tildef}
    L(\tildef, \pig)
    \,,\quad L(\tildef, \pig)
    &\triangleq
    \E_{d_{\pie}}
    [\log D_{\tildef, \pig}(s,a)]
    +
    \E_{d_{\pig}}
    [\log (1-D_{\tildef, \pig}(s,a))],\\
    \text{with}\quad D_{\tildef, \pig} &= \frac{\exp \tildef(s, a)}{\exp \tildef(s, a) + \pig(a|s)},
\end{aligned}
\end{equation}
for which we present the following theorem.
\begin{thm}
\label{thm:ASQF_thm}
For any generator policy $\pig$, the optimal discriminator parameter for \Eqref{eq:ASQF_obj} is 
\begin{equation*}
   \tildef^* \triangleq \argmax_{\tildef} L(\tildef, \pig) = \log\left(\pie(a|s)\frac{d_{\pie}(s)}{d_{\pig}(s)}\right) \, \forall (s,a) \in \gS \times \gA 
\end{equation*}
Using $\tildef^*$, the optimal generator policy $\pig^*$ is
\begin{align*} \argmin_{\pig}\max_{\tildef}L(\tildef, \pig) = \argmin_{\pig}L(\tildef^*, \pig) = \pie(a|s) = 
 \frac{\exp \tildef^*(s,a)}{\sum_{a'}\exp \tildef^*(s,a')} \,\, \forall (s,a) \in \gS\times \gA.
\end{align*}
\end{thm}
\begin{proof}
The beginning of the proof closely follows the proof of Appendix~\ref{app:proof_of_optim_D_gan}.
\begin{equation}
\begin{aligned}
    \argmax_{\tildef} L &(\tildef,\pig)= \\& \argmax_{\tildef} \sum_{s_i, a_i} d_{\pie}(s_i,a_i) \log D_{\tildef, \pig}(s_i,a_i) + d_{\pig}(s_i,a_i)\log(1-D_{\tildef, \pig}(s_i,a_i))
\end{aligned}
\end{equation}
We solve for each individual $(s_i,a_i)$ pair and note that $\tildef_{i} \mapsto D_i = \dfrac{\exp \tildef_{i}}{\exp \tildef_{i} + \pigi}$ is strictly monotonic on $\tildef_{i} \in \sR \, \forall \, \pigi \in ]0,1[$ so,
\begin{align}
    D^*_i=\argmax_{D_i} L_i
    \, \Leftrightarrow& \, \tildef^*_i =\argmax_{\tildef} L_i
\end{align}
Taking the derivative and setting it to 0, we find that
\begin{align}
 \left.\frac{d L_i}{d \tildef_i}\right|_{\tildef_i} = 0 \quad \Leftrightarrow \quad
    \tildef_i =\log\left(\pigi\frac{d_{\pie,i}}{d_{\pig,i}}\right)
\end{align}
We confirm that we have a global maximum with the second derivative test and the values at the border of the domain i.e. $\left.\dfrac{d^2 L_i}{d \tildef_i^2}\right|_{\tildef_i^*} < 0$ and $L_i$ goes to $-\infty$ for $\tildef_i \rightarrow +\infty$ and for $\tildef_i \rightarrow -\infty$.

It follows that
\begin{align}
    \tildef^*(s,a) &=\log\left(\pig(a|s)\frac{d_{\pie}(s,a)}{d_{\pig}(s,a)}\right) \quad \forall (s,a) \in \gS \times \gA\\
    \implies \tildef^*(s,a) &=\log\left(\cancel{\pig(a|s)}\frac{d_{\pie}(s)\pie(a|s)}{d_{\pig}(s)\cancel{\pig(a|s)}}\right) \quad \forall (s,a) \in \gS \times \gA\\
    \label{eq:ASQF_optimal_parameter}
    \implies \tildef^*(s,a) &=\log\left(\pie(a|s)\frac{d_{\pie}(s)}{d_{\pig}(s)}\right) \quad \forall (s,a) \in \gS \times \gA
\end{align}
This proves the first part of \Thmref{thm:ASQF_thm}.

To prove the second part notice that
\begin{equation}
\begin{aligned}
    D_{\tildef^*, \pig}(s,a) &= \dfrac{\pie(a|s)\dfrac{d_{\pie}(s)}{d_{\pig}(s)}}{\pie(a|s)\dfrac{d_{\pie}(s)}{d_{\pig}(s)} + \pig(a|s)}\\
    &= \frac{\pie(a|s)d_{\pie}(s)}{\pie(a|s)d_{\pie}(s) + \pig(a|s)d_{\pig}(s)}\\
    &= \frac{d_{\pie}(s,a)}{d_{\pie}(s,a) + d_{\pig}(s,a)}
\end{aligned}
\end{equation}
This is equal to the optimal discriminator of the GAN objective \Eqref{eq:D_optim_GAN_proof} when $x=(s,a)$. For this discriminator we showed in \Secref{app:proof_of_optim_D_gan} that the optimal generator $\pig^*$ is such that $d_{\pig^*}(s,a) = d_{\pie}(s,a)$ $\forall (s,a) \in \gS \times \gA$, which is satisfied for $\pig^*(a|s) = \pie(a|s)$ $\forall (s,a) \in \gS\times\gA$. 
Using the fact that
\begin{equation}
\label{eq:ASQF_partition_function}
    \sum_{a'}\exp\tildef^*(s,a') = \sum_{a'}\pie(a'|s)\frac{d_{\pie}(s)}{d_{\pig}(s)} = \frac{d_{\pie}(s)}{d_{\pig}(s)}\sum_{a'}\pie(a'|s) = \frac{d_{\pie}(s)}{d_{\pig}(s)}.
\end{equation}
we can combine \Eqref{eq:ASQF_optimal_parameter} and \Eqref{eq:ASQF_partition_function} to write the expert's policy $\pie$ as a function of the optimal discriminator parameter $\tildef^*$:
\begin{equation}
    \pie(a|s) = 
 \frac{\exp \tildef^*(s,a)}{\sum_{a'}\exp \tildef^*(s,a')} \,\, \forall (s,a) \in \gS\times \gA.
\end{equation}
This concludes the second part of the proof.
\end{proof}

\paragraph{Adversarial Soft-Q Fitting (ASQF) - practical algorithm}
In a nutshell, \Thmref{thm:ASQF_thm} tells us that training the discriminator in \Eqref{eq:ASQF_obj} to distinguish between transitions from the expert and transitions from a generator policy can be seen as retrieving $\tildef^*$ which plays the role of the expert's soft Q-function (i.e. which matches \Eqref{eq:max_ent_policy} for $\tildef^*=\frac{1}{\alpha}Q_{\text{soft},E}^*$):
\begin{equation}
\label{eq:ASQF_pi}
    \pie(a|s) = \frac{\exp \tildef^*(s,a)}{\sum_{a'}\exp \tildef^*(s,a')} = \exp\left(\tildef^*(s,a) - \log\sum_{a'}\exp \tildef^*(s,a') \right),
\end{equation}
Therefore, by training the discriminator, one simultaneously retrieves the optimal generator policy.

There is one caveat though: the summation over actions that is required in \Eqref{eq:ASQF_pi} to go from $\tildef^*$ to the policy is intractable in continuous action spaces and would require an additional step such as a projection to a proper distribution (\citet{haarnoja2018soft} use a Gaussian) in order to draw samples and evaluate likelihoods. Updating in this way the generator policy to match a softmax over our learned state-action preferences ($\tildef^*$) becomes very similar in requirements and computational load to a policy optimization step, thus defeating the purpose of this work which is to get rid of the policy optimization step. For this reason we only consider ASQF for discrete action spaces. 

As explained in \Secref{sec:asaf_practical_algorithm}, in practice we optimize $D_{\tildef, \pig}$ only for a few steps before updating $\pig$ by normalizing $\exp \tildef(s,a)$ over the action dimension. See Algorithm~\ref{alg:ASQF_alg} for the pseudo-code. 

\begin{algorithm}[H]
\label{alg:ASQF_alg}
  \begin{algorithmic}
  \caption{Adversarial Soft-Q Fitting (ASQF)}\label{alg:asqf}
    \REQUIRE expert transitions $\mathcal{D}_E = \{(s_i, a_i)\}_{i=1}^{N_E}$
    \STATE Randomly initialize $\tildef$ and get $\pig$ from \Eqref{eq:ASQF_pi}
    \FOR{steps $m=0$ to $M$}
      \STATE Collect transitions $\mathcal{D}_G = \{(s_i, a_i)\}_{i=1}^{N_G}$ by executing $\pig$
      \STATE Train $D_{\tildef, \pig}$ using binary cross-entropy on minibatches of transitions from $\mathcal{D}_E$ and $\mathcal{D}_G$
      \STATE Get $\pig$ from \Eqref{eq:ASQF_pi}
  \ENDFOR
  \end{algorithmic}
\end{algorithm}

\begin{figure}[h]
    \centering
    \includegraphics[width=.75\textwidth]{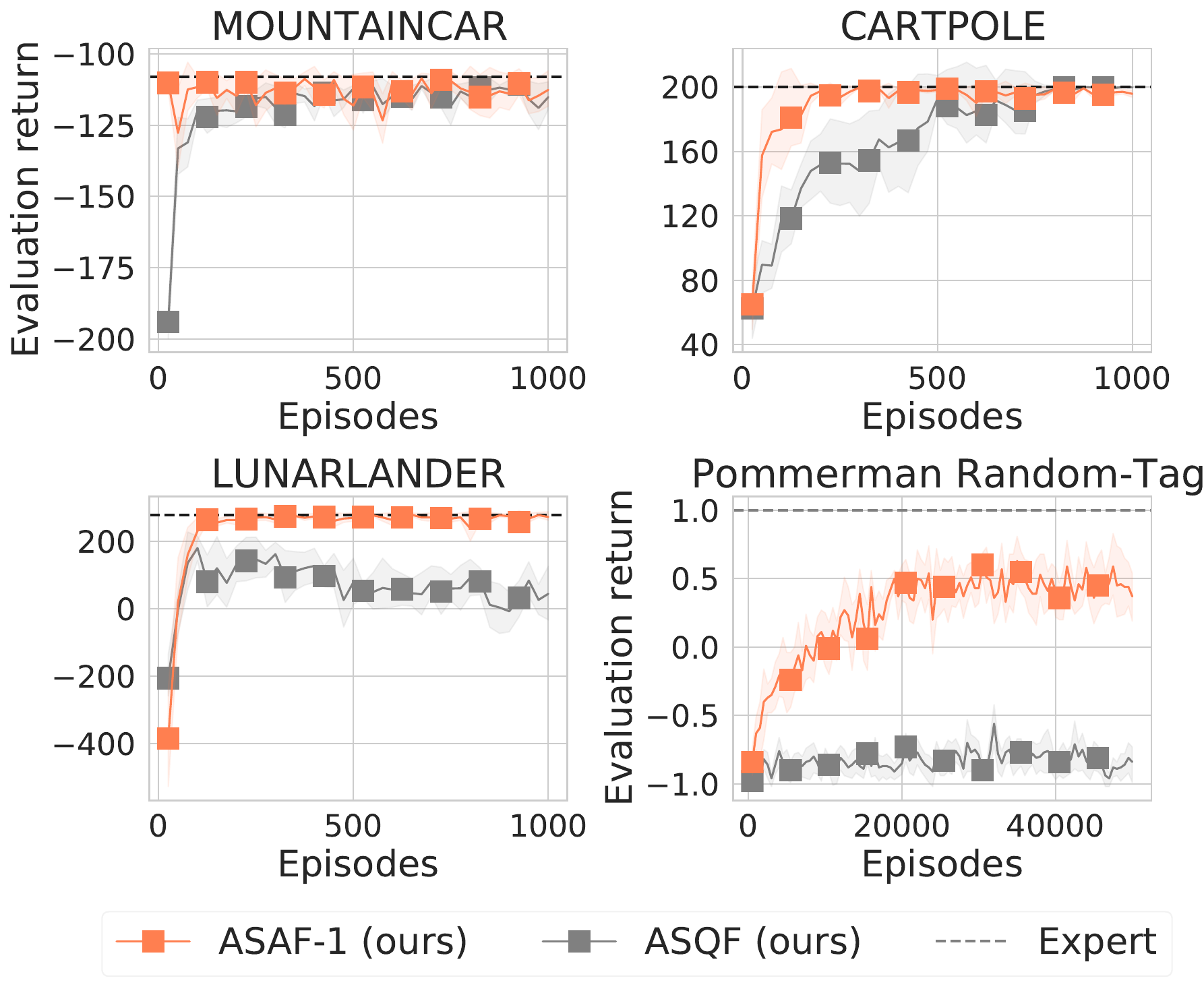}
    \caption{Comparison between ASAF-1 and ASQF, our two transition-wise methods, on environments with increasing observation space dimensionality}
    \label{fig:toy_asaf1_asqf}
\end{figure}

\paragraph{Experimental results}
\Figref{fig:toy_asaf1_asqf} shows that ASQF performs well on small scale environments but struggles and eventually fails on more complicated environments. Specifically, it seems that ASQF does not scale well with the observation space size. Indeed mountaincar, cartpole, lunarlander and pommerman have respectively an observation space dimensionality of 2, 4, 8 and 960. This may be due to the fact that the partition function \Eqref{eq:ASQF_partition_function} becomes more difficult to learn. Indeed, for each state, several transitions with different actions are required in order to learn it. Poorly approximating this partition function could lead to assigning too low a probability to expert-like actions and eventually failing to behave appropriately. ASAF on the other hand explicitly learns the probability of an action given the state -- in other word it explicitly learns the partition function -- and is therefore immune to that problem. 
\clearpage
\newpage
\section{Additional Experiments}
\label{app:additional_experiments}
\subsection{GAIL - Importance of Gradient Penalty}
\begin{figure}[h]
    \centering
    \includegraphics[width=1.\textwidth]{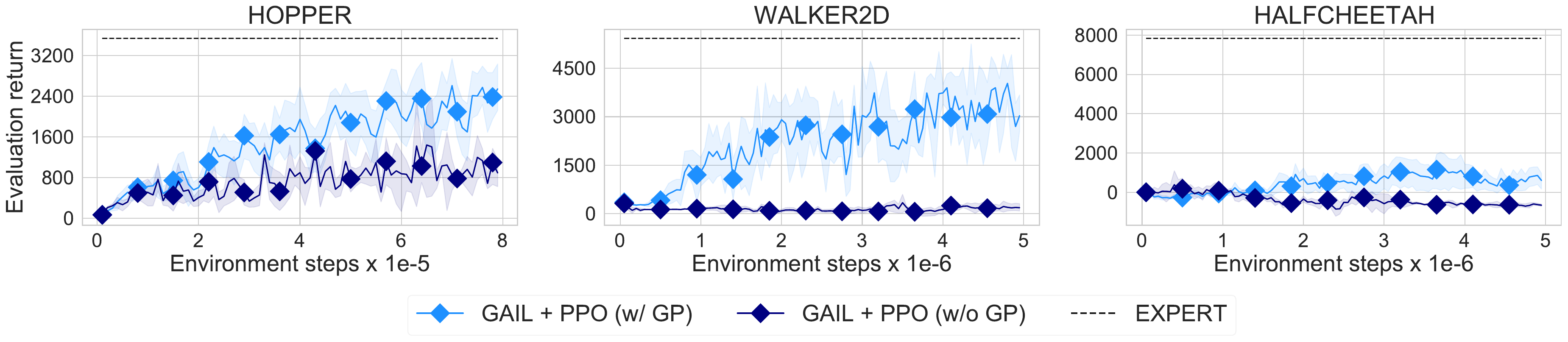}
    \caption{Comparison between original GAIL \cite{ho2016generative} and GAIL with gradient penalty (GP) \cite{gulrajani2017improved,kostrikov2018discriminatoractorcritic}}
    \label{fig:gail_gradient_penalty}
\end{figure}

\subsection{Mimicking the expert}
\noindent
\begin{minipage}[r]{0.5\textwidth}
To ensure that our method actually mimics the expert and doesn't just learn a policy that collects high rewards when trained with expert demonstrations, we ran ASAF-1 on the Ant-v2 MuJoCo environment using various sets of 25 demonstrations. These demonstrations were generated from a Soft Actor-Critic agent at various levels of performance during its training.  Since at low-levels of performance the variance of episode's return is high, we filtered collected demonstrations to lie in the targeted range of performance (e.g. return in [800, 1200] for the 1K set). Results in \Figref{fig:gradual_expert} show that our algorithm succeeds at learning a policy that closely emulates various demonstrators (even when non-optimal).
\end{minipage}\hfill
\begin{minipage}[l]{0.45\textwidth}
    \includegraphics[width=1.\textwidth]{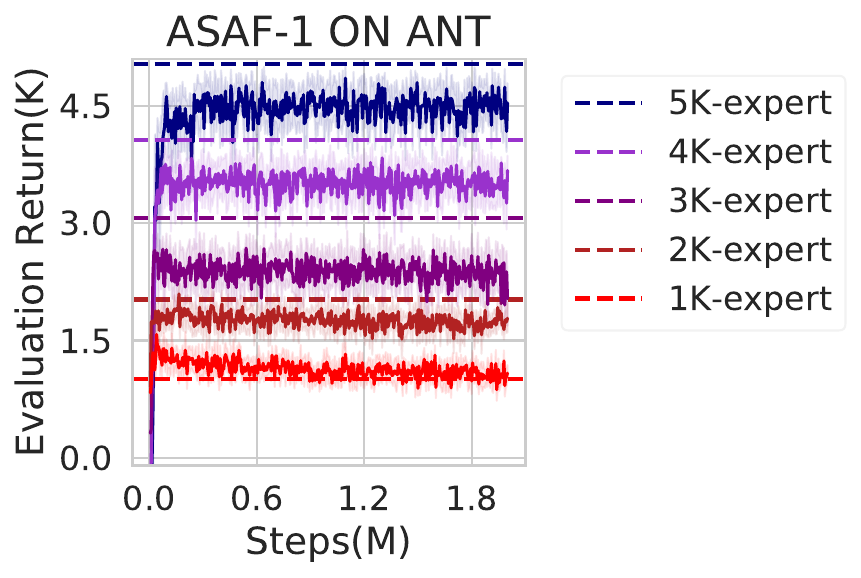}
    \captionof{figure}{ASAF-1 on Ant-v2. Colors are 1K, 2K, 3K, 4K, 5K expert's performance.}
    \label{fig:gradual_expert}
    \end{minipage}
    
\subsection{Wall Clock Time}
\begin{figure}[h]
    \centering
    \includegraphics[width=1.\textwidth]{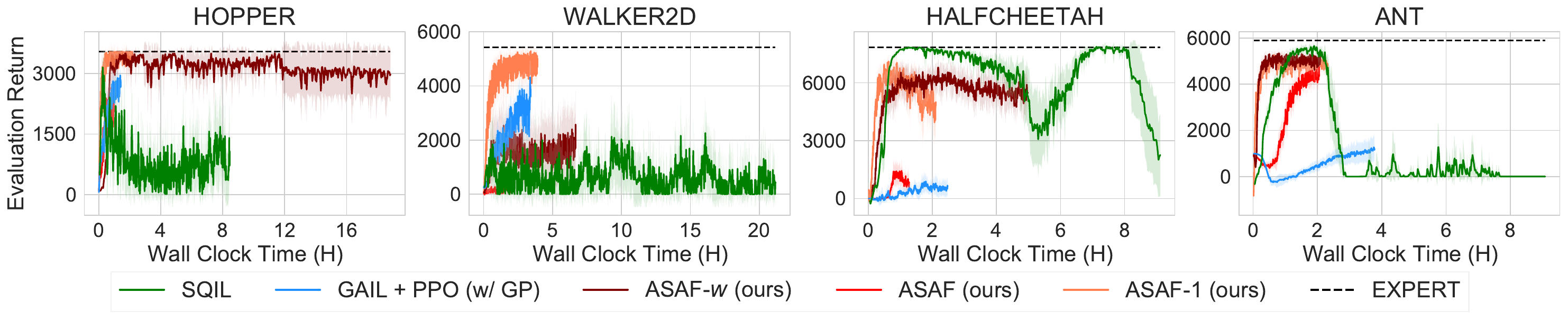}
    \caption{Training times on MuJoCo tasks for 25 expert demonstrations.}
    \label{fig:wall_clock_times}
\end{figure}

We report training times in \Figref{fig:wall_clock_times} and observe that ASAF-1 is always fastest to learn. Note however that reports of performance w.r.t wall-clock time should always be taken with a grain of salt as they are greatly influenced by hyper-parameters and implementation details.

\newpage
\section{Hyperparameter tuning and best configurations}
\label{app:hyperparameters}

\subsection{Classic Control}

For this first set of experiments, we use the fixed hyperparameters presented in Table~\ref{table:fixed_hyperparams_classic_control}.

\begin{table}[h]
\centering
\begin{sc}
\small
\caption{Fixed Hyperparameters for classic control tasks}
\begin{tabular}{lcc}
\label{table:fixed_hyperparams_classic_control}
\textbf{RL component}
\\ \hline
Hyper-parameter                                         &Discrete Control       &Continuous Control
\\ \hline
\\
\hspace{5mm} \textbf{SAC} \\
\hspace{5mm} Batch size (in transitions)                &256                    &256 \\
\hspace{5mm} Replay Buffer length $|\mathcal{B}|$       &$10^{6}$               &$10^{6}$ \\
\hspace{5mm} Warmup (in transitions)       &1280                      &10240 \\
\hspace{5mm} Initial entropy weight $\alpha$            &0.4                    &0.4 \\
\hspace{5mm} Gradient norm clipping threshold           &0.2                    &1 \\
\hspace{5mm} Transitions between update                 &40                     &1 \\
\hspace{5mm} Target network weight $\tau$               &0.01                   &0.01 \\
\\
\hspace{5mm} \textbf{PPO} \\
\hspace{5mm} Batch size (in transitions)                &256                    &256 \\
\hspace{5mm} GAE parameter $\lambda$                    &0.95                   &0.95 \\
\hspace{5mm} Transitions between update                 &-                      &2000 \\
\hspace{5mm} Episodes between updates                   &10                     &- \\
\hspace{5mm} Epochs per update                          &10                     &10 \\
\hspace{5mm} Update clipping parameter                  &0.2                    &0.2 \\
\\
\textbf{Reward Learning component}
\\ \hline
Hyper-parameter                                         &Discrete Control       &Continuous Control
\\ \hline
\\
\hspace{5mm} \textbf{AIRL, GAIL, ASAF-1} \\
\hspace{5mm} Batch size (in transitions)                &256                    &256 \\
\hspace{5mm} Transitions between update                 &-                      &2000 \\
\hspace{5mm} Episodes between updates                   &10                     &- \\
\hspace{5mm} Epochs per update                          &50                     &50 \\
\hspace{5mm} Gradient value clipping threshold         &-                      &1 \\
\hspace{5mm}(ASAF-1)&&\\
\\
\hspace{5mm} \textbf{ASAF, ASAF-\textit{w}} \\
\hspace{5mm} Batch size (in trajectories)               &10                     &10 \\
\hspace{5mm} Episodes between updates                   &10                     &20 \\
\hspace{5mm} Epochs per update                          &50                     &50 \\
\hspace{5mm} Window size $w$                            &(searched)             &200 \\
\hspace{5mm} Gradient value clipping threshold           &-                      &1 \\
\end{tabular}
\end{sc}
\end{table}

For the most sensitive hyperparameters, the learning rates for the reinforcement learning and discriminator updates ($\epsilon_{\text{RL}}$ and $\epsilon_{\text{D}}$), we perform a random search over 50 configurations and 3 seeds each (for each algorithm on each task) for 500 episodes. We consider logarithmic ranges, i.e. $\epsilon = 10^{u}$ with $u \sim Uniform(-6, -1)$ for $\epsilon_{\text{D}}$ and $u \sim Uniform(-4, -1)$ for $\epsilon_{\text{RL}}$. We also include in this search the critic learning rate coefficient $\kappa$ for PPO also sampled according to a logarithmic scale with $u \sim Uniform(-2, 2)$ so that the effective learning rate for PPO's critic network is $\kappa \cdot \epsilon_{\text{RL}}$. For discrete action tasks, the window-size \textit{w} for ASAF-\textit{w} is sampled uniformly within $\{32, 64, 128\}$. The best configuration for each algorithm is presented in Tables~\ref{table:hyperparams_cartpole}~to~\ref{table:hyperparams_lunarlander_c}. Figure~\ref{fig:results_toy} uses these configurations retrained on 10 seeds and twice as long.

Finally for all neural networks (policies and discriminators) for these experiments we use a fully-connected MLP with two hidden layers and ReLU activation (except for the last layer). We used hidden sizes of 64 for the discrete tasks and of 256 for the continuous tasks.

\begin{table}[h]
\centering
\begin{sc}
\caption{Best found hyper-parameters for Cartpole}
\resizebox{\textwidth}{!}{%
\begin{tabular}{lcccccc}
\label{table:hyperparams_cartpole}
Hyper-parameter                                     &ASAF           &ASAF-\textit{w}           &ASAF-$1$           &SQIL           &AIRL + PPO         &GAIL + PPO    
\\ \hline
Discriminator update lr $\epsilon_{\text{D}}$       &0.028          &0.039              &0.00046            &-              &2.5*$10^{-6}$        &0.00036 \\
RL update lr $\epsilon_{\text{RL}}$                 &-              &-                  &-                  &0.0067         &0.0052             &0.012 \\
Critic lr coefficient $\kappa$                      &-              &-                  &-                  &-              &0.25               &0.29 \\
window size \textit{w}                                     &-              &64                 &1                  &-              &-                  &- \\
window stride                                     &-              &64                 &1                  &-              &-                  &- \\
\end{tabular}}
\end{sc}
\end{table}

\begin{table}[h]
\centering
\begin{sc}
\caption{Best found hyper-parameters for Mountaincar}
\resizebox{\textwidth}{!}{%
\begin{tabular}{lcccccc}
\label{table:hyperparams_mountaincar}
Hyper-parameter                                     &ASAF           &ASAF-\textit{w}           &ASAF-$1$           &SQIL           &AIRL + PPO         &GAIL + PPO    
\\ \hline
Discriminator update lr $\epsilon_{\text{D}}$       &0.059          &0.059              &0.0088             &-              &0.0042             &0.00016 \\
RL update lr $\epsilon_{\text{RL}}$                 &-              &-                  &-                  &0.062          &0.016              &0.0022 \\
Critic lr coefficient $\kappa$                      &-              &-                  &-                  &-              &4.6                &0.018 \\
window size \textit{w}                                     &-              &32                 &1                  &-              &-                  &- \\
window stride                                     &-              &32                 &1                  &-              &-                  &- \\
\end{tabular}}
\end{sc}
\end{table}

\begin{table}[h]
\centering
\begin{sc}
\caption{Best found hyper-parameters for Lunarlander}
\resizebox{\textwidth}{!}{%
\begin{tabular}{lcccccc}
\label{table:hyperparams_lunarlander}
Hyper-parameter                                     &ASAF           &ASAF-\textit{w}           &ASAF-$1$           &SQIL           &AIRL + PPO         &GAIL + PPO    
\\ \hline
Discriminator update lr $\epsilon_{\text{D}}$       &0.0055         &0.0015             &0.00045            &-              &0.0002             &0.00019 \\
RL update lr $\epsilon_{\text{RL}}$                 &-              &-                  &-                  &0.0036         &0.0012             &0.0016 \\
Critic lr coefficient $\kappa$                      &-              &-                  &-                  &-              &0.48               &8.5 \\
window size \textit{w}                                     &-              &32                 &1                  &-              &-                  &- \\
window stride                                  &-              &32                 &1                  &-              &-                  &- \\
\end{tabular}}
\end{sc}
\end{table}

\begin{table}[h]
\centering
\begin{sc}
\caption{Best found hyper-parameters for Pendulum}
\resizebox{\textwidth}{!}{%
\begin{tabular}{lcccccc}
\label{table:hyperparams_pendulum}
Hyper-parameter                                     &ASAF           &ASAF-\textit{w}       &ASAF-$1$           &SQIL           &AIRL + PPO         &GAIL + PPO    
\\ \hline
Discriminator update lr $\epsilon_{\text{D}}$       &0.00069        &0.00082        &0.00046            &-              &4.3*$10^{-6}$      &1.6*$10^{-5}$ \\
RL update lr $\epsilon_{\text{RL}}$                 &-              &-              &-                  &0.0001         &0.00038            &0.00028 \\
Critic lr coefficient $\kappa$                      &-              &-              &-                  &-              &0.028              &84 \\
window size \textit{w}                              &-              &200                 &1                  &-              &-                  &- \\
window stride                                       &-              &200                 &1                  &-              &-                  &- \\
\end{tabular}}
\end{sc}
\end{table}

\begin{table}[h]
\centering
\begin{sc}
\caption{Best found hyper-parameters for Mountaincar-c}
\resizebox{\textwidth}{!}{%
\begin{tabular}{lcccccc}
\label{table:hyperparams_mountaincar_c}
Hyper-parameter                                     &ASAF           &ASAF-$w$           &ASAF-$1$           &SQIL           &AIRL + PPO         &GAIL + PPO    
\\ \hline
Discriminator update lr $\epsilon_{\text{D}}$       &0.00021        &3.8*$10^{-5}$      &6.2*$10^{-6}$      &-              &1.7*$10^{-5}$      &1.5*$10^{-5}$ \\
RL update lr $\epsilon_{\text{RL}}$                 &-              &-                  &-                  &0.0079         &0.0012             &0.0052 \\
Critic lr coefficient $\kappa$                      &-              &-                  &-                  &-              &10                 &12 \\
window size \textit{w}                              &-              &200                 &1                  &-              &-                  &- \\
window stride                                       &-              &200                 &1                  &-              &-                  &- \\
\end{tabular}}
\end{sc}
\end{table}

\begin{table}[h]
\centering
\begin{sc}
\caption{Best found hyper-parameters for Lunarlander-c}
\resizebox{\textwidth}{!}{%
\begin{tabular}{lcccccc}
\label{table:hyperparams_lunarlander_c}
Hyper-parameter                                     &ASAF           &ASAF-$w$           &ASAF-$1$           &SQIL           &AIRL + PPO         &GAIL + PPO    
\\ \hline
Discriminator update lr $\epsilon_{\text{D}}$       &0.0051         &0.0022             &0.0003             &-              &0.0045             &0.00014 \\
RL update lr $\epsilon_{\text{RL}}$                 &-              &-                  &-                  &0.0027         &0.00031            &0.00049 \\
Critic lr coefficient $\kappa$                      &-              &-                  &-                  &-              &14                 &0.01 \\
window size \textit{w}                              &-              &200                 &-                  &-              &-                  &- \\
window stride                                       &-              &200                 &-                  &-              &-                  &- \\
\end{tabular}}
\end{sc}
\end{table}

\clearpage
\subsection{MuJoCo}

For MuJoCo experiments (Hopper-v2, Walker2d-v2, HalfCheetah-v2, Ant-v2), the fixed hyperparameters are presented in Table~\ref{table:fixed_hyperparams_MuJoCo}. For all exeperiments, fully-connected MLPs with two hidden layers and ReLU activation (except for the last layer) were used, where the number of hidden units is equal to 256. 
\begin{table}[h]
\centering
\begin{sc}
\caption{Fixed hyperparameters for MuJoCo environments.}
\begin{tabular}{lcccc}
\label{table:fixed_hyperparams_MuJoCo}
\textbf{RL component}
\\ \hline
Hyper-parameter                                         &Hopper, Walker2d, HalfCheetah, Ant
\\ \hline
\\
\hspace{5mm} \textbf{PPO (for GAIL)} \\
\hspace{5mm} GAE parameter $\lambda$                    &{0.98} \\
\hspace{5mm} Transitions between updates                &{2000} \\
\hspace{5mm} Epochs per update                          &{5} \\
\hspace{5mm} Update clipping parameter                  &{0.2} \\
\hspace{5mm} Critic lr coefficient $\kappa$             &{0.25} \\
\hspace{5mm} Discount factor $\gamma$                          &{0.99} \\
\textbf{Reward Learning component}
\\ \hline
Hyper-parameter                                         &Hopper, Walker2d, HalfCheetah, Ant
\\ \hline
\\
\hspace{5mm} \textbf{GAIL} \\
\hspace{5mm} Transitions between updates                &{2000} \\
\\
\hspace{5mm} \textbf{ASAF}\\
\hspace{5mm} Episodes between updates                   &{25} \\
\\
\hspace{5mm} \textbf{ASAF-1 and ASAF-\textit{w}}\\
\hspace{5mm} Transitions between updates                &{2000} \\
\end{tabular}
\end{sc}
\end{table}

For SQIL we used SAC with the same hyperparameters that were used to generate the expert demonstrations. For ASAF, ASAF-1 and ASAF-\textit{w}, we set the learning rate for the discriminator at 0.001 and ran random searches over 25 randomly sampled configurations and 2 seeds for each task to select the other hyperparameters for the discriminator training. These hyperparameters included the discriminator batch size sampled from a uniform distribution over $\{10, 20, 30\}$ for ASAF and ASAF-\textit{w} (in trajectories) and over $\{100, 500, 1000, 2000\}$ for ASAF-1 (in transitions), the number of epochs per update sampled from a uniform distribution over $\{10, 20, 50\}$, the gradient norm clipping threshold sampled form a uniform distribution over $\{1, 10\}$, the window-size (for ASAF-\textit{w}) sampled from a uniform distribution over $\{100, 200, 500, 1000\}$ and the window stride (for ASAF-\textit{w}) sampled from a uniform distribution over $\{1, 50, w\}$. For GAIL, we obtained poor results using the original hyperparameters from \cite{ho2016generative} for a number of tasks so we ran random searches over 100 randomly sampled configurations for each task and 2 seeds to select for the following hyperparameters: the log learning rate of the RL update and the discriminator update separately sampled from uniform distributions over $[-7, -1]$, the gradient norm clipping for the RL update and the discriminator update separately sampled from uniform distributions over $\{None, 1, 10\}$, the number of epochs per update sampled from a uniform distribution over $\{5, 10, 30, 50\}$, the gradient penalty coefficient sampled from a uniform distribution over $\{1, 10\}$ and the batch size for the RL update and discriminator update separately sampled from uniform distributions over $\{100, 200, 500, 1000, 2000\}$. 

\begin{table}[h!]
\centering
\begin{sc}
\caption{Best found hyper-parameters for the Hopper-v2 environment}
\resizebox{\textwidth}{!}{%
\begin{tabular}{lccccccc}
\label{table:hopper}
Hyper-parameter                                 &ASAF           &ASAF-\textit{w}           &ASAF-$1$           &SQIL           &GAIL + PPO  
\\ \hline
RL batch size (in transitions)       &-              &-              &-         &256     &200 \\
Discriminator batch size (in transitions)       &-              &-              &100         &-     &2000 \\
Discriminator batch size (in trajectories)      &10              &10              &-         &-     &- \\
Gradient clipping (RL update)      &-              &-              &-         &-     &1. \\
Gradient clipping (discriminator update)  &10.              &10.              &1.         &-     &1. \\
Epochs per update &50              &50              &30         &-     &5 \\
Gradient penalty (discriminator update) &-              &-              &-         &-     &1. \\
RL update lr $\epsilon_{\text{RL}}$       &-     &-         &-         &$3*10^{-4}$     &$1.8*10^{-5}$        \\
Discriminator update lr $\epsilon_{\text{D}}$        &0.001     &0.001         &0.001         &-     &0.011         \\
window size \textit{w}       &-     &200         &1        &-     &-        \\
window stride       &-     &1         &1        &-     &-        \\
\end{tabular}}
\end{sc}
\end{table}

\begin{table}[h!]
\centering
\begin{sc}
\caption{Best found hyper-parameters for the HalfCheetah-v2 environment}
\resizebox{\textwidth}{!}{%
\begin{tabular}{lccccccc}
\label{table:halfcheetah}
Hyper-parameter                                 &ASAF           &ASAF-\textit{w}           &ASAF-$1$           &SQIL           &GAIL + PPO  
\\ \hline
RL batch size (in transitions)       &-              &-              &-         & 256    &1000 \\
Discriminator batch size (in transitions)       &-              &-              &100         &-     &100 \\
Discriminator batch size (in trajectories)      &10              &10              &-         &-     &- \\
Gradient clipping (RL update)      &-              &-              &-         &-     &- \\
Gradient clipping (discriminator update)  &10.              &1              &1         &-     &10 \\
Epochs per update &50              &10              &30         &-     &30 \\
Gradient penalty (discriminator update) &-              &-              &-         &-     &1. \\
RL update lr $\epsilon_{\text{RL}}$       &-     &-         &-         &$3*10^{-4}$     &0.0006        \\
Discriminator update lr $\epsilon_{\text{D}}$        &0.001     &0.001         &0.001         &-     &0.023         \\
window size \textit{w}       &-     &200         &1         &-     &-        \\
window stride       &-     &1         &1        &-     &-        \\
\end{tabular}}
\end{sc}
\end{table}

\begin{table}[h!]
\centering
\begin{sc}
\caption{Best found hyper-parameters for the Walker2d-v2 environment}
\resizebox{\textwidth}{!}{%
\begin{tabular}{lccccccc}
\label{table:walker2D}
Hyper-parameter                                 &ASAF           &ASAF-\textit{w}           &ASAF-$1$           &SQIL           &GAIL + PPO  
\\ \hline
RL batch size (in transitions)       &-              &-              &-         & 256    &200 \\
Discriminator batch size (in transitions)       &-              &-              &500         &-     &2000 \\
Discriminator batch size (in trajectories)      &20              &20              &-         &-     &- \\
Gradient clipping (RL update)      &-              &-              &-         &-     &- \\
Gradient clipping (discriminator update)  &10.              &1.              &10.         &-     &- \\
Epochs per update &30              &10              &50         &-     &30 \\
Gradient penalty (discriminator update) &-              &-              &-         &-     &1. \\
RL update lr $\epsilon_{\text{RL}}$       &-     &-         &-         &$3*10^{-4}$     &0.00039        \\
Discriminator update lr $\epsilon_{\text{D}}$        &0.001     &0.001         &0.001         &-     &0.00066         \\
window size \textit{w}       &-     &100         &1         &-     &-        \\
window stride       &-     &1         &1        &-     &-        \\
\end{tabular}}
\end{sc}
\end{table}

\begin{table}[h!]
\centering
\begin{sc}
\caption{Best found hyper-parameters for the Ant-v2 environment}
\resizebox{\textwidth}{!}{%
\begin{tabular}{lccccccc}
\label{table:ant}
Hyper-parameter                                 &ASAF           &ASAF-\textit{w}           &ASAF-$1$           &SQIL           &GAIL + PPO  
\\ \hline
RL batch size (in transitions)       &-              &-              &-         &256     &500 \\
Discriminator batch size (in transitions)       &-              &-              &100         &-     &100 \\
Discriminator batch size (in trajectories)      &20              &20              &-         &-     &- \\
Gradient clipping (RL update)      &-              &-              &-         &-     &- \\
Gradient clipping (discriminator update)  &10.              &1.              &1.         &-     &10. \\
Epochs per update &50              &50              &10         &-     &50 \\
Gradient penalty (discriminator update) &-              &-              &-         &-     &10 \\
RL update lr $\epsilon_{\text{RL}}$       &-     &-         &-         &$3*10^{-4}$     &$8.5*10^{-5}$        \\
Discriminator update lr $\epsilon_{\text{D}}$        &0.001     &0.001         &0.001         &-     &0.0016         \\
window size \textit{w}       &-     &200         &1         &-     &-        \\
window stride       &-     &50         &1        &-     &-        \\
\end{tabular}}
\end{sc}
\end{table}

\newpage
\subsection{Pommerman}

For this set of experiments, we use a number of fixed hyperparameters for all algorithms either inspired from their original papers for the baselines or selected through preliminary searches. These fixed hyperparameters are presented in Table~\ref{table:fixed_hyperparams_pommerman_random_tag}.

\begin{table}[h!]
\centering
\begin{sc}
\caption{Fixed Hyperparameters for Pommerman Random-Tag environment.}
\begin{tabular}{lcc}
\label{table:fixed_hyperparams_pommerman_random_tag}
\textbf{RL component}
\\ \hline
Hyper-parameter                                         &Pommerman Random-Tag
\\ \hline
\\
\hspace{5mm} \textbf{SAC} \\
\hspace{5mm} Batch size (in transitions)                &256 \\
\hspace{5mm} Replay Buffer length $|\mathcal{B}|$       &$10^{5}$ \\
\hspace{5mm} Warmup (in transitions)       &1280 \\
\hspace{5mm} Initial entropy weight $\alpha$            &0.4 \\
\hspace{5mm} Gradient norm clipping threshold           &0.2 \\
\hspace{5mm} Transitions between update                 &10 \\
\hspace{5mm} Target network weight $\tau$               &0.05 \\
\\
\hspace{5mm} \textbf{PPO} \\
\hspace{5mm} Batch size (in transitions)                &256 \\
\hspace{5mm} GAE parameter $\lambda$                    &0.95 \\
\hspace{5mm} Episodes between updates                   &10 \\
\hspace{5mm} Epochs per update                          &10 \\
\hspace{5mm} Update clipping parameter                  &0.2 \\
\hspace{5mm} Critic lr coefficient $\kappa$             &0.5 \\
\\
\textbf{Reward Learning component}
\\ \hline
Hyper-parameter                                         &Pommerman Random-Tag
\\ \hline
\\
\hspace{5mm} \textbf{AIRL, GAIL, ASAF-1} \\
\hspace{5mm} Batch size (in transitions)                &256 \\
\hspace{5mm} Episodes between updates                   &10 \\
\hspace{5mm} Epochs per update                          &10 \\
\\
\hspace{5mm} \textbf{ASAF, ASAF-\textit{w}} \\
\hspace{5mm} Batch size (in trajectories)               &5 \\
\hspace{5mm} Episodes between updates                   &10 \\
\hspace{5mm} Epochs per update                          &10 \\
\end{tabular}
\end{sc}
\end{table}

For the most sensitive hyperparameters, the learning rates for the reinforcement learning and discriminator updates ($\epsilon_{\text{RL}}$ and $\epsilon_{\text{D}}$), we perform a random search over 25 configurations and 2 seeds each for all algorithms. We consider logarithmic ranges, i.e. $\epsilon = 10^{u}$ with $u \sim Uniform(-7, -3)$ for $\epsilon_{\text{D}}$ and $u \sim Uniform(-4, -1)$ for $\epsilon_{\text{RL}}$. We also include in this search the window-size \textit{w} for ASAF-\textit{w}, sampled uniformly within $\{32, 64, 128\}$. The best configuration for each algorithm is presented in Table~\ref{table:hyperparams_pommerman}. Figure~\ref{fig:pommerman_results_randomTag} uses these configurations retrained on 10 seeds.

\begin{table}[h!]
\centering
\begin{sc}
\caption{Best found hyper-parameters for the Pommerman Random-Tag environment}
\resizebox{\textwidth}{!}{%
\begin{tabular}{lccccccc}
\label{table:hyperparams_pommerman}
Hyper-parameter         &ASAF           &ASAF-\textit{w}           &ASAF-$1$           &SQIL           &AIRL + PPO         &GAIL + PPO         &BC   
\\ \hline
Discriminator update lr $\epsilon_{\text{D}}$        &0.0007     &0.0002         &0.0001         &-     &3.1*$10^{-7}$         &9.3*$10^{-7}$         &0.00022\\
RL update lr $\epsilon_{\text{RL}}$       &-     &-         &-         &0.00019     &0.00017         &0.00015         &-\\
window size \textit{w}       &-     &32         &1         &-     &-         &-         &-\\
window stride       &-     &32         &1         &-     &-         &-         &-\\
\end{tabular}}
\end{sc}
\end{table}

Finally for all neural networks (policies and discriminators) we use the same architecture. Specifically, we first process the feature maps (see \Secref{app:env_pommerman}) using a 3-layers convolutional network with number of hidden feature maps of 16, 32 and 64 respectivelly. Each one of these layers use a kernel size of 3x3 with stride of 1, no padding and a ReLU activation. This module ends with a fully connected layer of hidden size 64 followed by a ReLU activation. The output vector is then concatenated to the unprocessed additional information vector (see \Secref{app:env_pommerman}) and passed through a final MLP with two hidden layers of size 64 and ReLU activations (except for the last layer).

\clearpage
\newpage
\section{Environments and expert data}
\label{app:environments}

\subsection{Classic Control}
The environments used here are the reference Gym implementations for classic control\footnote{See: \url{http://gym.openai.com/envs/\#classic_control}} and for Box2D\footnote{See: \url{http://gym.openai.com/envs/\#box2d}}.
We generated the expert trajectories for mountaincar (both discrete and continuous version) by hand using keyboard inputs. For the other tasks, we trained our SAC implementation to get experts on the discrete action tasks and our PPO implementation to get experts on the continuous action tasks.

\subsection{MuJoCo}
The experts were trained using our implementation of SAC~\cite{haarnoja2018soft} the state-of-the-art RL algorithm in MuJoCo continuous control tasks.
Our implementation basically refactors the SAC implementation from Rlpyt\footnote{See: 
\url{https://github.com/astooke/rlpyt}}. We trained SAC agent for 1,000,000 steps for Hopper-v2 and 3,000,000 steps for Walker2d-v2 and HalfCheetah-v2 and Ant-v2. 
We used the default hyper-parameters from Rlpyt.

\subsection{Pommerman}
\label{app:env_pommerman}
The observation space that we use for Pommerman domain~\cite{resnick2018pommerman} is composed of a set of 15 feature maps as well as an additional information vector. The feature maps whose dimensions are given by the size of the board (8x8 in the case of 1vs1 tasks) are one-hot across the third dimension and represent which element is present at which location. Specifically, these feature maps identify whether a given location is the current player, an ally, an ennemy, a passage, a wall, a wood, a bomb, a flame, fog, a power-up. Other feature maps contain integers indicating bomb blast stength, bomb life, bomb moving direction and flame life for each location. Finally, the additional information vecor contains the time-step, number of ammunition, whether the player can kick and blast strengh for the current player. The agent has an action space composed of six actions: do-nothing, up, down, left, right and lay bomb.

For these experiments, we generate the expert demonstrations using Agent47Agent, the open-source champion algorithm of the FFA 2018 competition~\cite{zhou2018hybrid} which uses hardcoded heuristics and Monte-Carlo Tree-Search\footnote{See: \url{https://github.com/YichenGong/Agent47Agent/tree/master/pommerman}}. While this agent occasionally eliminates itself during a match, we only select trajectories leading to a win as being expert demonstrations.

\subsection{Demonstrations summary}
Table~\ref{table:demo_summary} provides a summary of the expert data used.

\begin{table}[h]
\centering
\begin{sc}
\caption{Expert demonstrations used for Imitation Learning}
\resizebox{\textwidth}{!}{%
\begin{tabular}{lcc}
\label{table:demo_summary}
Task-Name         &Expert mean return            &Number of expert trajectories  
\\ \hline
Cartpole &200.0     &10 \\
Mountaincar & -108.0      &10\\
Lunarlander & 277.5      &10\\
Pendulum & -158.6     &10\\
Mountaincar-c & 93.92    &10\\
Lunarlander-c & 266.1    &10\\
Hopper &3537        &25\\
Walker2D &5434      &25\\
Halfcheetah &7841       &25\\
Ant &5776       &25\\
Pommerman random-tag        &1 &300, 150, 75, 15, 5, 1\\
\end{tabular} }
\end{sc}
\end{table}

\end{document}